\documentclass[11pt]{article}

\usepackage{acl}
\usepackage{times}
\usepackage{latexsym}
\usepackage[T1]{fontenc}
\usepackage[utf8]{inputenc}
\usepackage{microtype}
\usepackage{inconsolata}
\usepackage{natbib}
\usepackage{amsmath,amssymb,amsthm}
\usepackage{booktabs}
\usepackage{multirow}
\usepackage{array}
\usepackage{makecell}
\usepackage{graphicx}
\usepackage{float}
\usepackage{xcolor}
\usepackage{enumitem}
\usepackage{tikz}

\usetikzlibrary{arrows.meta,positioning}
\usepackage{pifont}

\newcommand{\cmark}{\ding{51}}
\newcommand{\xmark}{\ding{55}}
\newcommand{\pmark}{\ensuremath{\triangle}}
\newcolumntype{P}[1]{>{\raggedright\arraybackslash}p{#1}}

\newcommand\samethanks[1][\value{footnote}]{\footnotemark[#1]}

\title{DPA-MoE: A Runtime Precision Control Framework for MoE Inference}

\author{\makecell{
  Jaeyun Kim$^{1}$\thanks{\hspace{.06in}Equal contribution},
  Hoonki Lee$^{1}$\samethanks,
  Susie Shin$^{2}$, \\
  Hyunmin Kim$^{2}$,
  Dahyun Lee$^{2}$,
  Sangyeob Kim$^{2}$\thanks{\hspace{.06in}Corresponding author}} \\
$^{1}$Agency for Defense Development \quad\quad $^{2}$Yonsei University \\
\texttt{\{kimjy.greg, hoonki2002\}@gmail.com} \\
\texttt{\{susie3529, ryan3504, dahyunlee17, sangyeob.kim\}@yonsei.ac.kr}
}

\begin{document}
\maketitle

\begin{abstract}
Mixture-of-Experts (MoE) models reduce computation by routing each
token to a small set of experts, but each selected expert is still used
at a fixed precision. We present \textbf{DPA-MoE}, a runtime precision
control framework that makes routed experts any-precision. DPA-MoE
encodes expert feed-forward weights as prefix-decodable chunks using
SharedScale AnyBCQ, profiles per-expert reconstruction residuals
offline, and allocates chunks at inference time using live router gate
scores. This post-routing allocation lets a single encoded MoE
checkpoint operate at multiple effective weight-memory budgets without
retraining or re-encoding. Experiments on four MoE models spanning
different scales, routing widths, and shared-expert configurations show
that DPA-MoE consistently improves reconstruction fidelity over uniform
chunk allocation and yields a better overall weight-size--quality
frontier, with modest steady-state throughput overhead in our prototype.
\end{abstract}

\section{Introduction}

Large language models are used across increasingly diverse workloads,
including interactive assistants, coding tools, document summarization,
and retrieval-augmented question answering. These workloads vary in
their latency, throughput, memory, and quality requirements, creating a
need for multiple operating points from the same model. However, a
conventional fixed-precision checkpoint exposes only one point on this
resource--quality curve. Releasing or storing a separate quantized
checkpoint for each operating point increases storage and deployment
complexity. An attractive alternative is an
\emph{any-precision} model: a single encoded artifact whose effective
weight size can be adjusted at inference time~\citep{
yu2021anyprecisiondnn, park2024anyprecision, wang2025bitstack,
anybcq2025}.

The need is sharpest for local inference, where the binding constraints
change while the model is resident: memory shifts as other applications come and go and a user may want speed now and accuracy
next. Adapting a fixed-precision checkpoint means swapping in a
separately quantized artifact, which requires re-encoding or fetching
another multi-gigabyte file; an any-precision artifact instead moves
along the size--quality curve in place.

\begin{figure}[!t]
  \centering
  \includegraphics[width=\columnwidth]{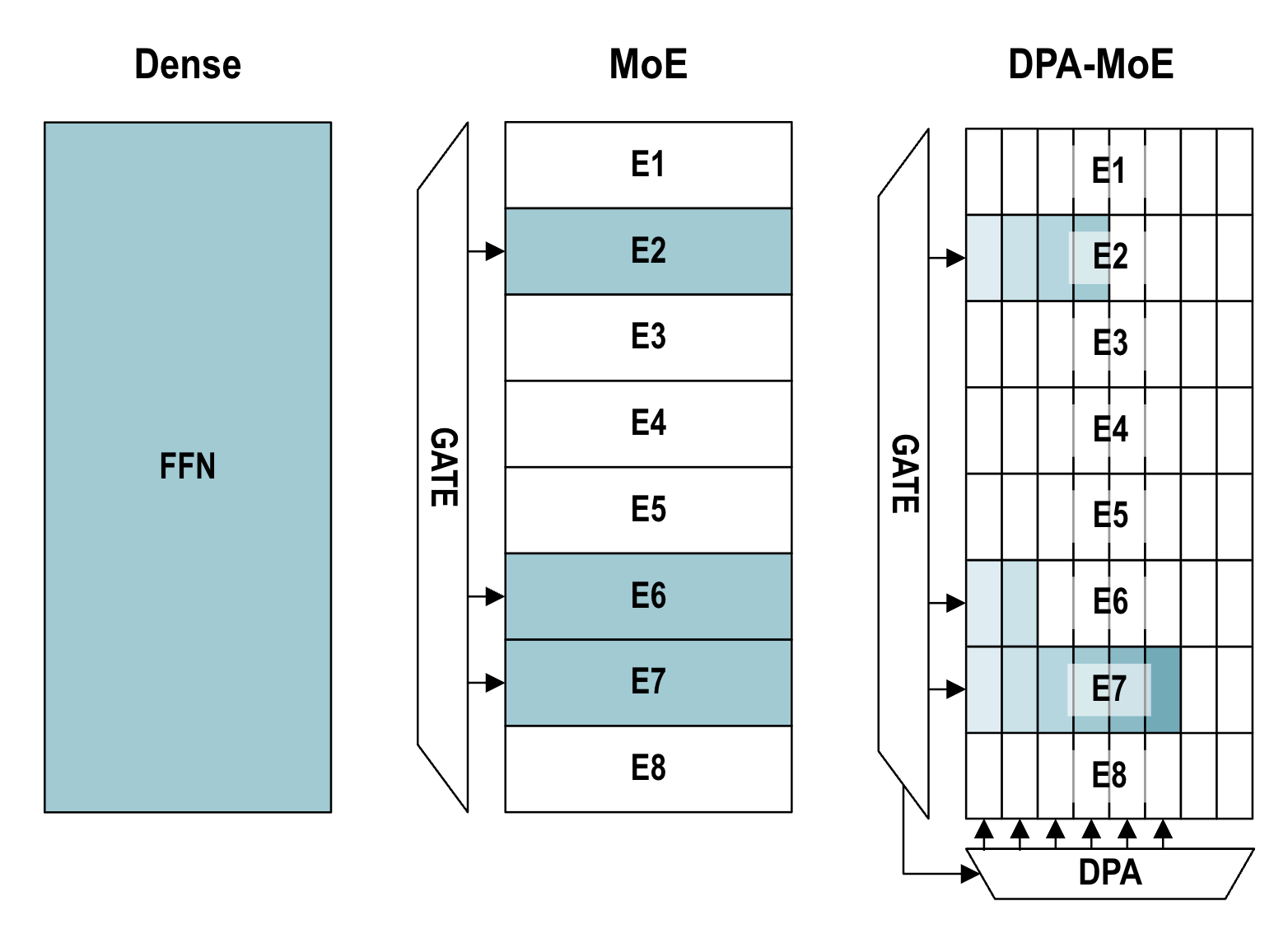}
  \caption{DPA-MoE adds post-routing precision allocation to MoE:
  routing chooses \emph{which} experts run, while DPA-MoE chooses
  \emph{how much} of each selected expert is used.}
  \label{fig:intro}
\end{figure}

Mixture-of-Experts (MoE) models route each token to a small subset of
experts rather than applying the full feed-forward parameter set~\citep{
shazeer2017outrageously, lepikhin2021gshard, fedus2022switch,
du2022glam, jiang2024mixtral, dai2024deepseekmoe, muennighoff2024olmoe}.
This increases total capacity while keeping per-token computation
small. However, MoE only controls \emph{which} experts are activated:
once selected, an expert is still evaluated at a fixed precision. Since
expert FFN weights dominate modern MoE parameter counts (about 95\% in
Qwen3-30B-A3B~\citep{qwen3team2025}), the number of weight bits read
inside each selected expert remains an important control point.

This paper makes selected experts any-precision. As shown in
Figure~\ref{fig:intro}, MoE chooses \emph{which} experts are activated,
while DPA-MoE chooses \emph{how much} of each selected expert is used by
assigning a variable number of precision chunks
(Section~\ref{sec:chunk-def}) per routed activation. Throughout,
\emph{precision} means this chunk count $k$ rather than a numeric
bit-width: an expert at precision $k$ is one reconstructed from its
first $k$ chunks.
Smaller chunk budgets give lighter but less accurate reconstructions;
larger budgets recover higher fidelity. This lets a single MoE
checkpoint trace a fine-grained weight-size--quality frontier without
retraining or maintaining multiple quantized variants.

A uniform any-precision MoE assigns the same chunk count to every
activated expert, ignoring router-provided structure. Existing methods
address memory and quality through post-training quantization~\citep{
frantar2023qmoe, duanmu2025mxmoe} or runtime adaptation using cache
state, device constraints, or long-horizon activation statistics~\citep{
tang2024hobbit, wang2025d2moe, chu2025dynaexq}. DPA-MoE instead
combines offline quantization residuals with live gate scores, since
experts differ both in current-token importance and reconstruction
sensitivity.

We present \textbf{DPA-MoE}, a dynamic precision allocation framework
for any-precision MoE models. DPA-MoE encodes expert FFN weights with
SharedScale AnyBCQ, a prefix-decodable representation built on
AnyBCQ~\citep{anybcq2025}, profiles reconstruction residuals for each
expert and chunk count, and combines those residuals with live router
gate scores at inference time. The pretrained router and expert
identities remain unchanged; DPA-MoE only controls how much precision
each selected expert receives. It is not an end-to-end serving system,
but a precision-allocation primitive complementary to systems such as
DeepSpeed-MoE, HOBBIT, and D2MoE~\citep{rajbhandari2022deepspeedmoe,
tang2024hobbit, wang2025d2moe}.

On Qwen3-30B-A3B, DeepSeekMoE-16B-Chat, OLMoE-1B-7B-0125, and
Qwen1.5-MoE-A2.7B \citep{qwen3team2025, dai2024deepseekmoe,
muennighoff2024olmoe, qwen15moe}, DPA-MoE consistently reduces KL
divergence relative to uniform chunk allocation and yields a better
overall weight-size--quality frontier, with modest steady-state
throughput overhead in our prototype. Sparse routing chooses the
experts; DPA-MoE chooses how much of each selected expert to use.

Our contributions are:
\begin{itemize}
    \item We formulate \emph{any-precision MoE}, where sparse
    routing selects experts and a separate precision controller
    adjusts how many weight bits are read from each selected expert.

    \item We introduce SharedScale AnyBCQ, a prefix-decodable
    expert encoding built on AnyBCQ~\citep{anybcq2025} that
    lets a single MoE checkpoint expose multiple effective
    weight-memory budgets without retraining or re-encoding.

    \item We propose a two-stage allocation framework that 
    combines offline residual profiling with live router gate 
    scores to allocate precision chunks across routed expert 
    activations.

    \item We show that DPA-MoE improves the weight-size--quality
    frontier over uniform chunk allocation, while incurring 
    modest steady-state throughput overhead in a lightweight 
    prototype.
\end{itemize}

\section{Related Work}

\paragraph{Static Quantization.}
Post-training quantization reduces LLM memory footprint 
by compressing weights offline. General LLM post-training
quantization methods include LLM.int8~\citep{dettmers2022llmint8},
ZeroQuant~\citep{yao2022zeroquant}, SmoothQuant~\citep{xiao2023smoothquant},
GPTQ~\citep{frantar2023gptq}, and AWQ~\citep{lin2024awq}.
MoE-specific methods extend this to expert 
weights, either uniformly~\citep{frantar2023qmoe} or with 
mixed precision assigned per expert based on sensitivity 
and activation frequency~\citep{duanmu2025mxmoe}.
While mixed-precision methods acknowledge that experts 
differ in importance, the allocation is fixed before 
deployment and remains constant regardless of the 
current input.

\paragraph{Input-Adaptive Precision Methods.}
Recent work has begun to adapt expert precision beyond 
static assignment.
\citet{tang2024hobbit} serve cache-miss experts at lower 
precision, but the decision is binary and driven by cache 
state rather than token-level routing confidence.
\citet{chu2025dynaexq} maintain a high-precision resident 
set based on long-horizon router traces, but rely on 
aggregate statistics that cannot reflect the importance 
of the current token to each expert.
\citet{wang2025d2moe} insert a per-expert bit-width gate 
downstream of the unmodified router, but require training 
an additional adapter and bind precision to a learned 
policy rather than exposing it as a separable runtime 
variable.

\paragraph{Progressive and Multi-Precision Encodings.}
A complementary line of work encodes weights such that 
multiple precisions are accessible without re-encoding.
\citet{yu2021anyprecisiondnn} introduce neural networks whose
precision can be changed at runtime, \citet{park2024anyprecision} 
show that a single quantized 
model can support multiple bit-widths, \citet{wang2025bitstack}
provide fine-grained model size control through stacked residual
quantization, and \citet{anybcq2025} advance this through progressive 
bitplanes with negligible switching overhead.
These encodings provide the representational substrate
for runtime precision control, but do not specify how
that control should be exercised across experts.

\paragraph{Positioning.}
Appendix~\ref{app:method-comparison} compares these methods on five
axes: training-free, router left unchanged, token-aware, fine-grained
budgets reachable, and no re-fetch on a precision upgrade. The closest
is D\textsuperscript{2}MoE, whose nested Matryoshka quantization reuses
already-loaded bits as our prefix-decodable encoding does, but it
exposes only discrete tiers and needs a trained per-expert network to
choose bit-widths. DPA-MoE is, to our knowledge, the only method that is
simultaneously training-free, token-aware, and fine-grained in $k$ while
leaving the router untouched.

\section{Background and Problem Formulation}
\subsection{Precision Chunks}
\label{sec:chunk-def}

DPA-MoE stores each expert weight tensor as an ordered sequence of
\emph{chunks}. A chunk is a single binary bitplane together with its
per-group scale; a per-group offset is shared across chunks. Using the
first $k$ chunks appends bitplanes to a prefix that is already valid, so
raising $k$ never invalidates what has been read; $k_{\min}$ and
$k_{\max}$ denote the smallest and largest supported chunk counts.
Section~\ref{sec:encoding} gives the reconstruction
(Equation~\ref{eq:bcq-reconstruction}) and the encoding procedure. A
chunk is thus the granularity at which expert precision can be spent.

\subsection{MoE Routing}

For token state $x$, an MoE layer selects experts $\mathcal{E}(x)$ and
outputs $y=\sum_{e\in\mathcal{E}(x)}g_e(x)E_e(x)$. We assume each expert
can support prefix reconstruction with chunk count $k$, where larger
$k$ uses more of the encoded expert weight and generally gives a more
faithful reconstruction.

\subsection{Problem Statement}
Uniform precision is structurally mismatched with MoE routing. For a
given token, the router does not activate experts equally: it assigns
different gate scores to different selected experts, indicating that
their contributions to the layer output are not symmetric. Across
layers and experts, the reconstruction error introduced by using a
shorter precision prefix also varies. Nevertheless, a uniform precision
policy reads the same number of weight bits from every selected expert.

This wastes budget in two ways. First, low-contribution experts can
receive the same precision as high-contribution experts. Second,
experts that are already easy to reconstruct can receive the same
precision as experts whose output is more sensitive to missing chunks.
As a result, uniform precision can spend memory and bandwidth on
expert activations with small marginal value while under-serving the
activations where additional precision matters most.

The problem addressed in this paper is therefore not to change the MoE
router, retrain the model, or design a new serving scheduler. It is to
remove this uniform-precision assumption inside the experts selected
by the existing router, so that a single encoded MoE checkpoint can
expose multiple weight-size--quality operating points more efficiently.

\section{Method}
\subsection{Progressive Expert Encoding}
\label{sec:encoding}

\begin{figure*}[t]
  \centering
  \includegraphics[width=\textwidth]{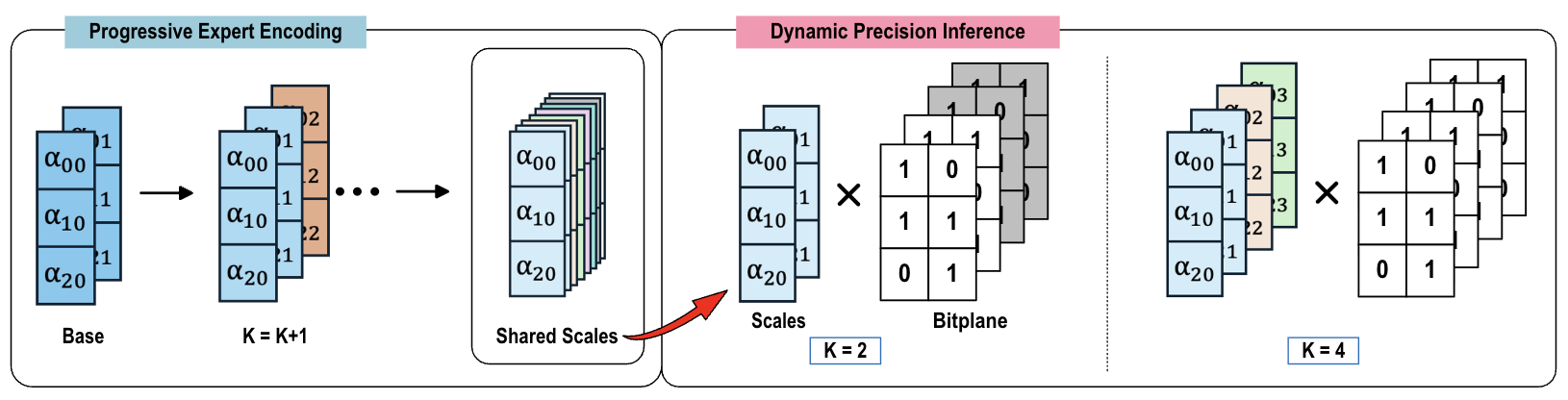}
  \caption{
    \textbf{SharedScale AnyBCQ encoding and dynamic precision inference.}
    \textit{Left (Progressive Expert Encoding):} chunks are added 
    sequentially, with all active $\alpha$ planes jointly re-optimized 
    at each step to produce a single shared scale table.
    \textit{Right (Dynamic Precision Inference):} at inference time, 
    selecting $k$ chunks loads the first $k$ bitplanes and scale 
    columns from the shared table --- no additional scale fetches 
    are required when switching precision.
  }
  \label{fig:encoding}
\end{figure*}

DPA-MoE encodes each expert feed-forward weight tensor as an ordered
sequence of precision chunks, where loading a longer prefix generally
improves reconstruction fidelity. Following AnyBCQ~\citep{anybcq2025},
each grouped weight tensor at precision $k$ is approximated as
\begin{equation}
W^{(k)} \approx \beta + \sum_{p=1}^{k} \alpha_p B_p,
\label{eq:bcq-reconstruction}
\end{equation}
where $B_p \in \{-1, +1\}^d$ is a binary bitplane, $\alpha_p$ is a
per-group scale, and $\beta$ is an optional per-group offset, with group size
$G{=}128$. The default precision range spans $k_{\min}{=}2$ to
$K{=}8$ chunks.

Standard AnyBCQ re-optimizes $\alpha_p$ independently at each precision 
tier, producing per-tier scale snapshots. Supporting the full 2--8 bit 
range therefore requires storing one scale set per tier, resulting in 
substantially larger per-expert storage overhead compared to a shared 
parameterization (Appendix~\ref{app:memory-analysis}).

We therefore introduce \textbf{SharedScale AnyBCQ}, which fixes a 
single shared $\alpha$ and $\beta$ table across all precision tiers 
during encoding. At inference time, moving from $k$ to $k{+}1$ chunks 
requires loading only the new bitplane $B_{k+1}$ and its scale 
$\alpha_{k+1}$; previously loaded scales remain valid and unchanged.
To compensate for the reduced per-tier flexibility, the encoding 
optimization objective is:
\begin{equation}
\mathcal{L} = \mathcal{L}_{\text{block}} +
\lambda \cdot \mathcal{L}_{\text{preserve}},
\label{eq:encoding-loss}
\end{equation}
where
\begin{equation}
\begin{aligned}
\mathcal{L}_{\text{block}} &=
\mathrm{MSE}\!\left(F_q(x_{\text{fp}}),\,
F_{\text{fp}}(x_{\text{fp}})\right), \\
\mathcal{L}_{\text{preserve}} &=
\mathbb{E}_{k \in [k_{\min}, K{-}1]}\left[
\mathrm{MSE}\!\left(W^{(k)},\; W_{\text{fp}}\right)\right],
\end{aligned}
\label{eq:encoding-loss-terms}
\end{equation}
where $\lambda$ is a weighting hyperparameter that balances 
reconstruction fidelity at full precision against quality preservation 
at intermediate precisions. Here $F_{\text{fp}}$ denotes layer $L$ 
evaluated with full-precision expert weights, and $F_q$ denotes the 
same layer with expert weights quantized at precision $K$, both 
evaluated on the same full-precision cached activation $x_{\text{fp}}$. 
This isolates the reconstruction error attributable to the current 
layer under calibration.
$\mathcal{L}_{\text{block}}$ minimizes the layer-level output 
error at full chunk count $K$, while $\mathcal{L}_{\text{preserve}}$ 
explicitly protects reconstruction quality at all intermediate 
precisions $k \in [k_{\min}, K{-}1]$.
Hyperparameters and calibration details are provided in 
Table~\ref{tab:encoding-config}.

Figure~\ref{fig:encoding} illustrates this process.
During encoding, chunks are added one at a time; at each step, 
all active $\alpha$ scales are jointly re-optimized via 
$\mathcal{L}$ with signs $B$ fixed, gradually refining the 
shared scale table to account for all intermediate precisions 
simultaneously. At inference time, selecting precision $k$ loads 
the first $k$ chunks, enabling precision switching without 
re-fetching previously loaded scales.

\subsection{Two-Stage Framework}
\label{sec:allocation}

\begin{figure*}[t]
  \centering
  \includegraphics[width=\textwidth, height=0.33\textheight, 
  keepaspectratio]{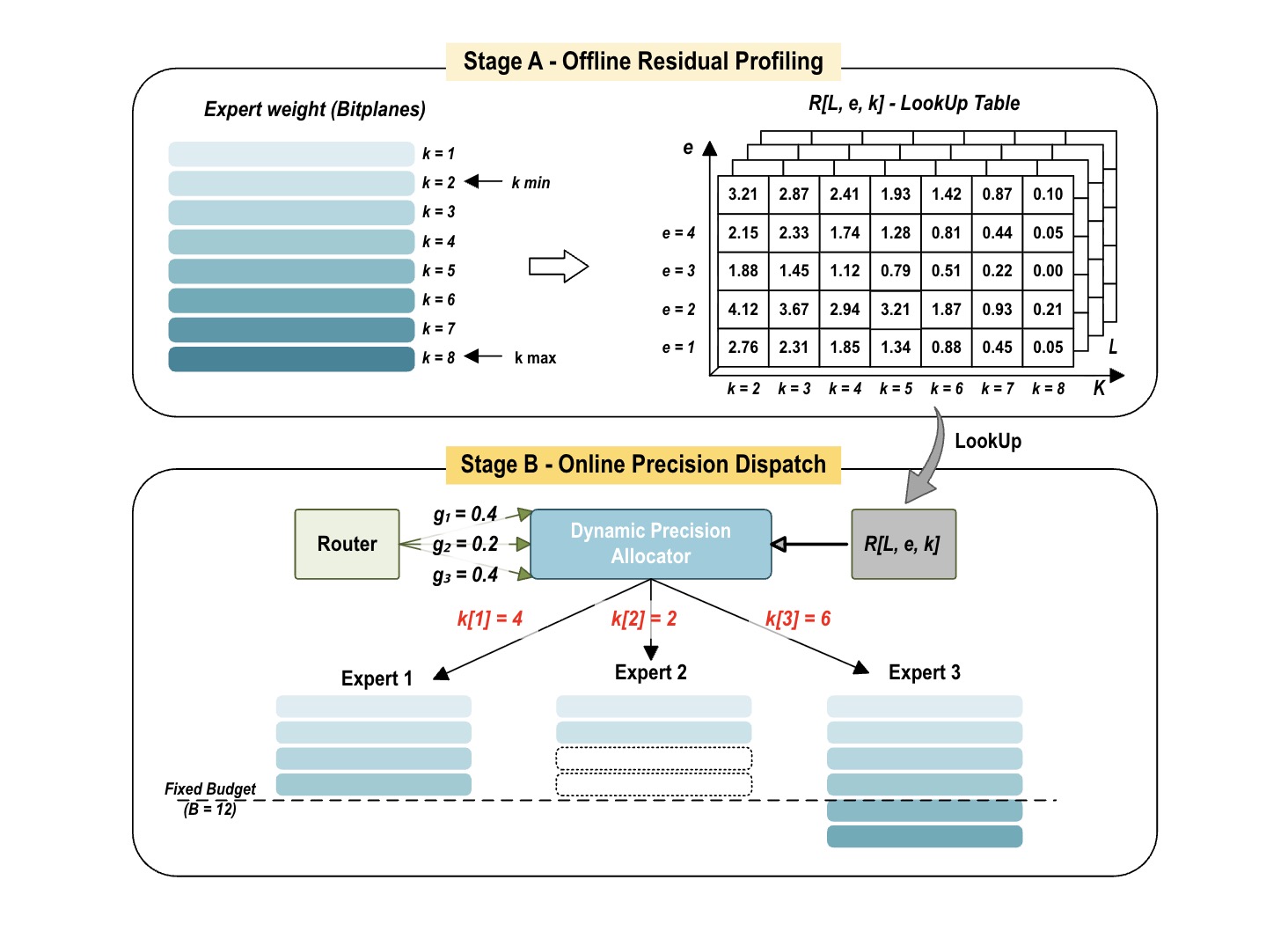}
  \caption{
    \textbf{Two-Stage Precision Allocation.} Offline profiling builds a residual table $\mathcal{R}[L,e,k]$. At inference time, DPA combines this table with live gate scores and allocates a fixed chunk budget across routed experts.
  }
  \label{fig:two-stage}
\end{figure*}

The encoding described in Section~\ref{sec:encoding} converts 
expert precision into a runtime-controllable variable. The central 
question then becomes: \textit{given a fixed per-dispatch budget, 
how should chunks be allocated across active experts to best 
preserve output quality?} A uniform allocation---assigning the 
same chunk count to every expert---ignores two sources of 
heterogeneity: the intrinsic quantization error of each expert, 
which varies across experts independently of routing, and the 
router's gate scores, which reflect each expert's relevance to 
the current token. We propose a two-stage framework that keeps
these two sources strictly separate.

\paragraph{Stage A: Offline Residual Profiling.}
Prior to inference, we measure the reconstruction error incurred by each expert \(e\) in layer \(L\) at every precision level \(k\).
With \(\mathcal{D}_L\) the input distribution to experts in layer \(L\), we define
\begin{align}
R[L,e,k]
=
\mathbb{E}_{x \sim \mathcal{D}_L}
\left[
\left\|
\Delta E_e^{(k)}(x)
\right\|^2
\right],
\end{align}
where \(\Delta E_e^{(k)}(x) = E_e(x) - E_e^{(k)}(x)\) is the difference between the full-precision and \(k\)-chunk reconstruction of expert \(e\).
In practice this expectation is approximated on a calibration set.
Profiling is gate-free: it captures only each expert's intrinsic compression error.
The result is a Residual LUT \(\mathcal{R} = \{R[L,e,k]\}\), computed once and stored alongside the model.

\paragraph{Stage B: Online Precision Allocation.}
At inference time, for each MoE dispatch, the router provides 
live gate scores $\{g_e\}$ for the $n$ active experts. Together
with the Residual LUT $\mathcal{R}$, these give the Dynamic
Precision Allocator~(DPA) everything it needs to choose $k_e$
per expert.

\paragraph{Objective Derivation.}
The two stages combine through the layer-level output error. For a
token $x$, let $\Delta Y=\sum_{e\in\mathcal{E}(x)}
g_e(x)\Delta E_e^{(k_e)}(x)$ denote the reconstruction error of the
MoE layer output. Expanding $\|\Delta Y\|^2$ gives diagonal terms
$\sum_e g_e(x)^2\|\Delta E_e^{(k_e)}(x)\|^2$ and cross terms
$\sum_{e\neq e'} g_e(x)g_{e'}(x)
\langle \Delta E_e^{(k_e)}(x),\Delta E_{e'}^{(k_{e'})}(x)\rangle$.
Dropping the cross terms gives

\begin{equation}
\|\Delta Y\|^2
\approx
\sum_e g_e(x)^2
\bigl\|\Delta E_e^{(k_e)}(x)\bigr\|^2 .
\label{eq:separable-error}
\end{equation}

Although this approximation changes the absolute magnitude of the
objective, it preserves allocation rankings: the exact
and approximated objectives have Spearman correlation at least $0.999$
across all precision levels (Appendix~\ref{app:cross-term}).

Taking the expectation over the calibration distribution yields
$g_e(x)^2 R[L,e,k_e]$ as the per-expert cost: a live per-dispatch
coefficient times the offline LUT. Before allocation, we apply a
monotone projection along $k$ to remove calibration noise, yielding
$\widetilde{R}$ with
$\widetilde{R}[L,e,k{+}1]\leq\widetilde{R}[L,e,k]$.

For batched prefill, we aggregate gate coefficients over
the microbatch $\mathcal{X}$:
\begin{equation}
G_e=\sum_{x\in\mathcal{X}:e\in\mathcal{E}(x)} g_e(x)^2 .
\label{eq:gate-aggregate}
\end{equation}
The same objective then applies with $g_e(x)^2$ replaced by $G_e$.

\subsection{Dynamic Precision Allocator (DPA)}
\label{sec:dp}

The DPA solves the following budgeted allocation problem at every MoE
dispatch:
\begin{align}
\min_{\{k_e\}} \sum_{e\in\mathcal{A}}
w_e\cdot \widetilde{R}[L,e,k_e]
\quad \text{s.t.} \quad
\sum_{e\in\mathcal{A}} k_e = B,
\label{eq:objective}
\end{align}
where $\mathcal{A}$ is the active expert set, $w_e=g_e(x)^2$ for
single-token decoding and $w_e=G_e$ for batched prefill. The budget is
$B=\mathrm{round}(|\mathcal{A}|\cdot\bar{k})$, and
$k_e\in\{k_{\min},\dots,k_{\max}\}$. For decoding,
$\mathcal{A}=\mathcal{E}(x)$; for batched prefill, $\mathcal{A}$ is the
union of experts activated by tokens in the microbatch.

\paragraph{Multi-Choice Knapsack DPA.}
Equation~\ref{eq:objective} is a multi-choice knapsack problem~\citep{
kellerer2004knapsack}. Let
$\mathcal{A}=\{e_1,\dots,e_n\}$ and define
$c_i(k)=w_{e_i}\widetilde{R}[L,e_i,k]$. Let $f(i,b)$ be the minimum
cost for the first $i$ experts under partial budget $b$:
\begin{equation}
\begin{aligned}
f(i,b)
&=
\min_{k}
\left[f(i-1,b-k)+c_i(k)\right],\\
&\quad k\in\{k_{\min},\dots,k_{\max}\}.
\end{aligned}
\label{eq:dp-recursion}
\end{equation}
with $f(0,0)=0$ and $f(0,b)=\infty$ for $b>0$. The optimum is
$f(n,B)$, and the selected chunk counts are recovered by backtracking.
Since the DP enumerates all feasible chunk assignments under the
budget, it returns the global optimum of
Equation~\ref{eq:objective} with respect to the projected residual
table $\widetilde{R}$.

\paragraph{Computational Cost.}
The DPA runs in
$O(n\cdot B\cdot k_{\mathrm{range}})$ time, where
$n=|\mathcal{A}|$, $B=\mathrm{round}(n\bar{k})$, and
$k_{\mathrm{range}}=k_{\max}-k_{\min}+1$. Equivalently, the cost is
$O(n^2\bar{k}k_{\mathrm{range}})$. For single-token decoding,
$n$ is the top-$k$ routing count; for Qwen3-30B-A3B,
$n=8$, $B\le64$, and $k_{\mathrm{range}}=7$, giving at most
$8\cdot64\cdot7=3{,}584$ candidate transitions per dispatch.

For batched prefill, $n$ can grow because $\mathcal{A}$ is the union of
experts selected by all tokens, but it remains bounded by the number of
experts in the layer rather than by the number of token--expert pairs.
In the worst case with $n\le128$, $\bar{k}\le8$, and
$k_{\mathrm{range}}=7$, the DP evaluates at most
$128\cdot(128\cdot8)\cdot7=917{,}504$ transitions per MoE layer
dispatch. This cost is amortized over the microbatch: for $m$ tokens,
the per-token cost scales as
$O(n^2\bar{k}k_{\mathrm{range}}/m)$ with
$n\le\min(N_{\mathrm{experts}},m k_{\mathrm{top}})$. Thus, small
batches have very small active sets, while large batches may activate
more experts but divide the allocator cost across more tokens.

\section{Experimental Setup}

\subsection{Models}
\label{sec:models}

We evaluate on four open-weight MoE language models, chosen to vary in
scale, routing width, and shared-expert configuration:
Qwen3-30B-A3B \citep{qwen3team2025}, 8 of 128 experts, no shared
experts; DeepSeekMoE-16B-Chat \citep{dai2024deepseekmoe}, 6 of 64 routed
experts plus 2 shared; OLMoE-1B-7B-0125 \citep{muennighoff2024olmoe},
8 of 64, no shared experts; and
Qwen1.5-MoE-A2.7B \citep{qwen15moe}, 4 of 60 routed experts plus 4
shared. The first two carry the full evaluation suite including
inference speed; the latter two are evaluated on KL divergence and
downstream accuracy.

\subsection{Baselines}
\label{sec:baselines}

We compare DPA against two baselines. As static
quantization references, we include Q2\_K, Q5\_K\_M,
and Q8\_0 GGUF quantizations of all four models,
representing fixed operating points at approximately
2-, 5-, and 8-bit compression. For Qwen3-30B-A3B and
DeepSeekMoE-16B-Chat we use published GGUF
artifacts \citep{unsloth_qwen3, mradermacher_deepseek}; the
OLMoE-1B-7B-0125 and Qwen1.5-MoE-A2.7B artifacts are converted
locally from the base checkpoints with llama.cpp.
As a dynamic baseline, Uniform $k$ serves all experts
at a fixed chunk count $k \in \{2, \dots, 8\}$,
using the same encoding as DPA but without
gate-aware allocation.

\subsection{Evaluation}
\label{sec:evaluation}

We evaluate reconstruction quality via mean
KL divergence (KLD) between the quantized and
full-precision output distributions, measured on
10 sequences of 4{,}096 tokens sampled from the
WikiText-2 test set~\citep{merity2017pointer} with
a fixed random seed. For downstream task performance,
we report accuracy on four benchmarks:
ARC-Challenge~\citep{clark2018think},
ARC-Easy~\citep{clark2018think},
WinoGrande~\citep{sakaguchi2021winogrande}, and
PIQA~\citep{bisk2020piqa}. We additionally report
the macro-average across all four tasks as a
single summary metric.

\subsection{Inference Evaluation Setup}
\label{sec:inference-setup}

We evaluate inference throughput and latency under a controlled
execution setting designed to compare the cost of materializing and
computing different amounts of expert weight. All methods use the same
model, implementation, and execution path; the only
difference is the chunk allocation assigned to activated experts.
In the Full-Memory setting used for the speed measurements, all encoded
expert chunks are resident in GPU memory, so the measurement isolates
precision allocation and variable-chunk execution rather than SSD,
host-cache, or expert-offloading effects.

We compare DPA against Uniform-$k$ at matched average chunk budgets
$\bar{k}\in\{3,4,5,6,7\}$. Uniform-$k$ materializes the same number of
chunks for every activated expert, while DPA redistributes the same
total chunk budget according to gate-weighted residual costs. Thus, the
comparison isolates the runtime and quality effect of non-uniform
expert precision allocation under the same effective weight-memory
budget.

We report generation throughput (TPS) and time-to-first-token (TTFT)
averaged over 100 sequences from each of WikiText-2~\citep{merity2017pointer},
C4-en~\citep{raffel2020t5}, and the Korean split of
mC4~\citep{xue2021mt5}.

\begin{figure*}[!t]
    \centering
    \begin{minipage}[t]{0.48\textwidth}
        \centering
        \includegraphics[width=\textwidth]{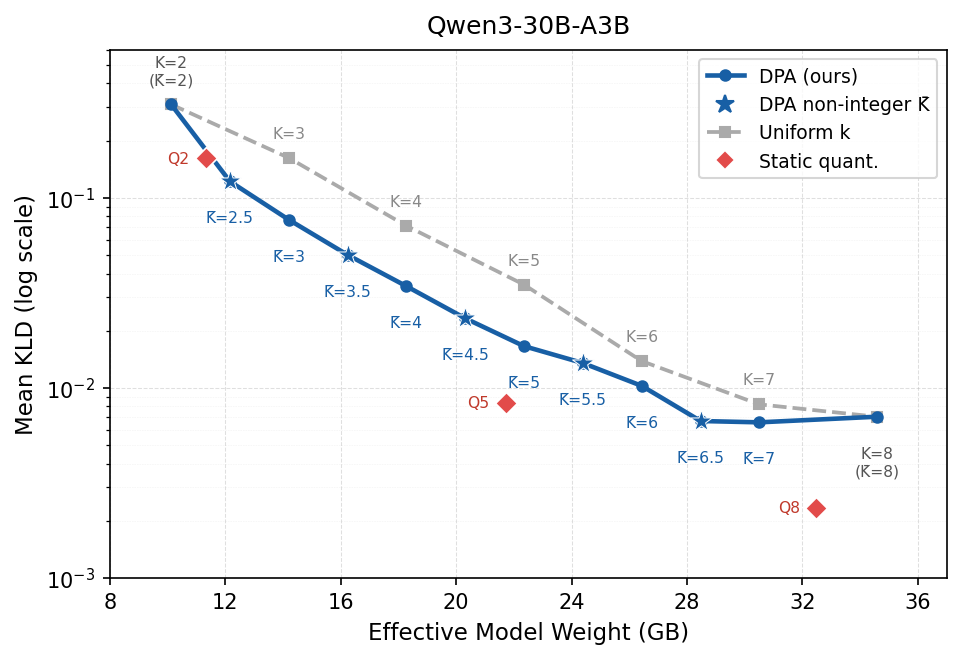}
    \end{minipage}
    \hfill
    \begin{minipage}[t]{0.48\textwidth}
        \centering
        \includegraphics[width=\textwidth]{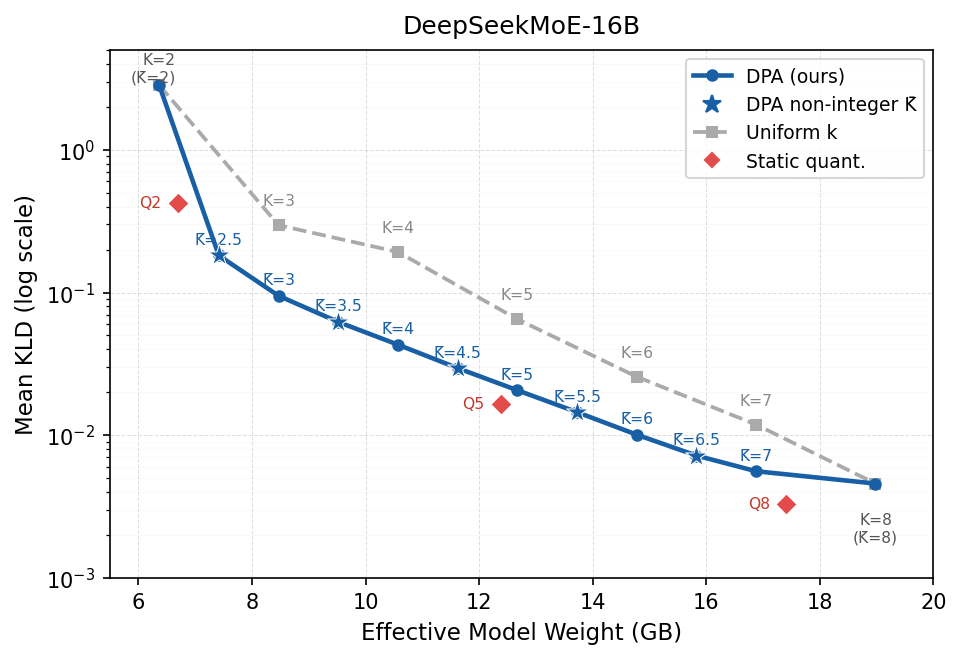}
    \end{minipage}
    \caption{%
        Mean KL divergence vs.\ effective model weight (GB) for DPA (ours),
        Uniform-$k$ allocation, and static quantization baselines
        on Qwen3-30B-A3B (left) and DeepSeekMoE-16B-Chat (right).
    }
    \label{fig:kld_comparison}
\end{figure*}

\begin{figure*}[!t]
    \centering
    \begin{minipage}[t]{0.48\textwidth}
        \centering
        \includegraphics[width=\textwidth]{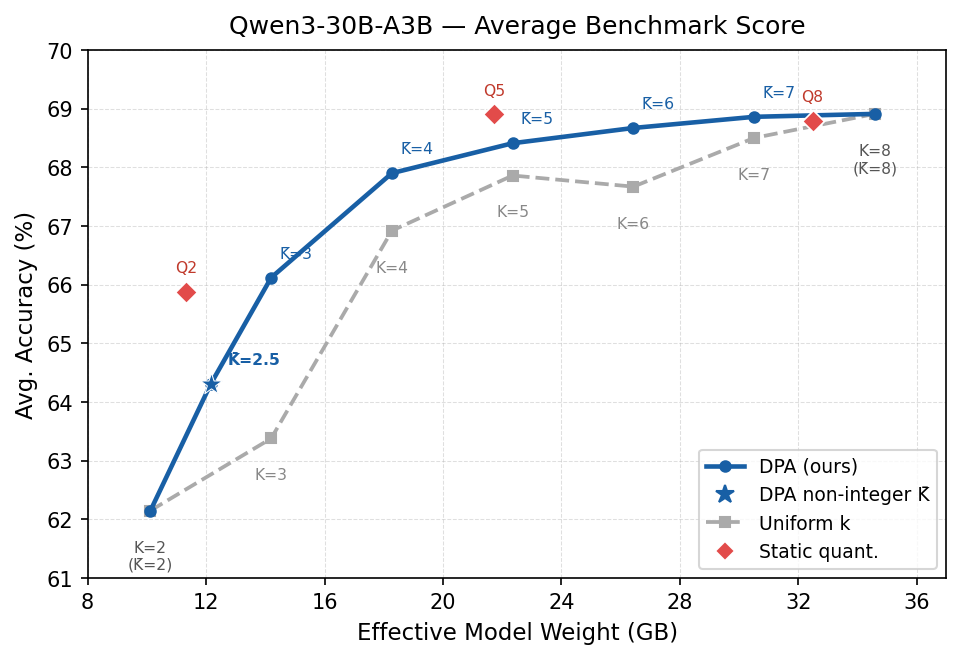}
    \end{minipage}
    \hfill
    \begin{minipage}[t]{0.48\textwidth}
        \centering
        \includegraphics[width=\textwidth]{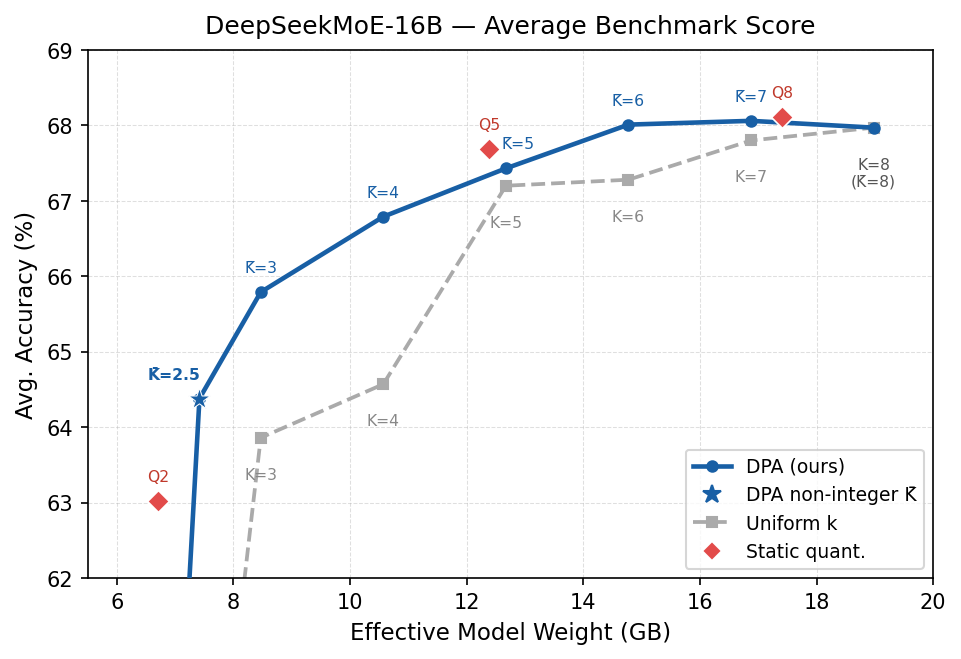}
    \end{minipage}
    \caption{%
        Macro-average benchmark accuracy (\%) vs.\ effective model weight (GB)
        for DPA (ours), Uniform-$k$ allocation, and static quantization baselines
        on Qwen3-30B-A3B (left) and DeepSeekMoE-16B-Chat (right).
    }
    \label{fig:avg_score_comparison}
\end{figure*}

\section{Results}

\subsection{Reconstruction Quality}
\label{sec:kld}

We adopt mean KL divergence (KLD) between the quantized
and full-precision output distributions as our primary
reconstruction metric.
Unlike perplexity, KLD directly quantifies how much
quantization distorts the model's predicted distribution
at every token, providing a sensitive and task-agnostic
measure of fidelity that captures both mode shifts and
tail deformations in the output.

Figure~\ref{fig:kld_comparison} plots mean KLD against
effective model weight for DPA and all baselines.
On both models, the DPA curve lies consistently below the
Uniform~$k$ curve across the full compression range,
confirming that gate-aware allocation reduces output
distortion at every matched memory budget.
The gain is most pronounced in the mid-compression
regime, where DPA reduces KLD by up to $4.5\times$
relative to Uniform~$k$ at matched size
(DeepSeekMoE-16B-Chat, $\bar{k}{=}4$),
and recovers equivalent fidelity at $2$--$4$~GB
less memory across both models.
Non-integer operating points (e.g.,\ $\bar{k}{=}2.5$)
smoothly interpolate between integer anchors,
enabling fine-grained memory--fidelity trade-offs
that static quantization cannot provide.

\subsection{Downstream Task Performance}
\label{sec:benchmark}

Figure~\ref{fig:avg_score_comparison} reports macro-average
accuracy across ARC-Challenge, ARC-Easy, WinoGrande, and PIQA
as a function of effective model weight;
per-task breakdowns are provided in Appendix~\ref{app:pertask}.

DPA consistently outperforms Uniform~$k$ at matched memory
budgets on both models, with the largest gains at aggressive
compression where uniform allocation is most wasteful.
The accuracy improvements mirror the KLD reductions reported
in Section~\ref{sec:kld}, confirming that reduced output
distortion translates to measurable downstream gains.

\subsection{Generality across MoE Architectures}
\label{sec:generality}

We repeat the same protocol on OLMoE-1B-7B-0125 and Qwen1.5-MoE-A2.7B,
which bracket the shared-expert axis --- OLMoE has none, Qwen1.5-

\begin{table*}[!h]
\centering
\small
\begin{tabular*}{\textwidth}{@{\extracolsep{\fill}}lcccccccccccc}
\toprule
& \multicolumn{6}{c}{\textbf{Qwen3-30B-A3B}}
& \multicolumn{6}{c}{\textbf{DeepSeekMoE-16B-Chat}} \\
\cmidrule(lr){2-7} \cmidrule(lr){8-13}
& \multicolumn{2}{c}{\textbf{Avg Acc (\%)}} 
& \multicolumn{2}{c}{\textbf{TPS}} 
& \multicolumn{2}{c}{\textbf{TTFT (s)}}
& \multicolumn{2}{c}{\textbf{Avg Acc (\%)}} 
& \multicolumn{2}{c}{\textbf{TPS}} 
& \multicolumn{2}{c}{\textbf{TTFT (s)}} \\
\cmidrule(lr){2-3}\cmidrule(lr){4-5}\cmidrule(lr){6-7}
\cmidrule(lr){8-9}\cmidrule(lr){10-11}\cmidrule(lr){12-13}
$\bar{k}$ & Uni & DPA & Uni & DPA & Uni & DPA 
          & Uni & DPA & Uni & DPA & Uni & DPA \\
\midrule
3 & 63.38 & 66.12 & 58.70 & 55.29 & 1.001 & 1.311
  & 63.86 & 65.79 & 116.54 & 106.96 & 0.519 & 0.691 \\
4 & 66.92 & 67.90 & 51.53 & 48.72 & 1.150 & 1.560
  & 64.57 & 66.79 & 105.68 & 97.06 & 0.597 & 0.842 \\
5 & 67.86 & 68.41 & 46.29 & 44.23 & 1.277 & 1.773
  & 67.20 & 67.43 & 95.17 & 89.64 & 0.679 & 0.978 \\
6 & 67.67 & 68.67 & 41.00 & 38.79 & 1.427 & 1.898
  & 67.28 & 68.01 & 88.68 & 82.18 & 0.758 & 1.101 \\
7 & 68.50 & 68.86 & 35.97 & 34.53 & 1.603 & 2.084
  & 67.80 & 68.06 & 82.05 & 76.68 & 0.801 & 1.203 \\
\bottomrule
\end{tabular*}
\caption{Macro-average benchmark accuracy, generation throughput
(TPS), and time-to-first-token (TTFT) on Qwen3-30B-A3B and
DeepSeekMoE-16B-Chat under the Full-Memory setting.}
\label{tab:tps}
\end{table*}
MoE has four --- and differ in routing width ($8$-of-$64$ versus $4$-of-$60$);
full sweeps are in Appendix~\ref{app:extra-models}. On both, DPA reduces
KL divergence relative to Uniform~$k$ by $26$--$69\%$ wherever the
allocation is not forced, and improves accuracy at most matched budgets:
the macro-average rises from $60.60\%$ to $62.40\%$ at $\bar{k}{=}3$ on
OLMoE, and from $62.04\%$ to $63.45\%$ at $\bar{k}{=}4$ on Qwen1.5-MoE.
The methods coincide at $\bar{k}{=}2$ and $\bar{k}{=}8$, where
$B=|\mathcal{A}|\bar{k}$ admits one feasible assignment. Across all four
models, the benefit of post-routing allocation is not confined to a
particular routing design.

\subsection{Inference Speed}
\label{sec:tps}

Table~\ref{tab:tps} reports macro-average accuracy, generation
throughput (TPS), and time-to-first-token (TTFT) for Uniform~$k$ and
DPA at matched $\bar{k}$ values under the Full-Memory setting.

At the same $\bar{k}$, DPA incurs a generation throughput overhead 
of $4$--$6\%$ on Qwen3-30B-A3B and $6$--$8\%$ on DeepSeekMoE-16B-Chat 
relative to Uniform~$k$, along with a TTFT increase of 
approximately $30$--$50\%$, both attributable to the per-dispatch 
knapsack allocation.

Because DPA consistently achieves higher accuracy at every $\bar{k}$, 
the more informative comparison is at matched quality. 
On Qwen3-30B-A3B, DPA at $\bar{k}{=}4$ achieves $67.90\%$ accuracy 
at $48.72$ TPS, nearly matching Uniform $k{=}5$ in accuracy 
($67.86\%$) while delivering approximately $5\%$ higher throughput 
than that Uniform setting ($46.29$ TPS).
On DeepSeekMoE-16B-Chat, DPA at $\bar{k}{=}4$ ($66.79\%$, $97.06$ TPS) 
similarly exceeds Uniform $k{=}5$ ($67.20\%$, $95.17$ TPS) in 
throughput while trailing by only $0.41$ accuracy points.
Across both models, DPA recovers the throughput cost of dynamic 
allocation by operating at a lower $\bar{k}$ than its uniform 
counterpart at comparable quality.

\subsection{Which Signal Drives the Gain?}
\label{sec:signal-ablation}

Beating Uniform-$k$ shows that non-uniform allocation helps, but not
that our objective is the right one. We therefore isolate the two
signals Equation~\ref{eq:objective} combines: \emph{Gate-only} replaces
the per-expert residual with a layer-wise average,
$g_e^2\cdot\bar{R}[L,k]$, and \emph{Residual-only} fixes the gate
coefficient to a constant, $1\cdot R[L,e,k]$. We report KLD, the metric
tied most directly to the allocator's objective. This stands in for a
head-to-head comparison against published methods, which we could not
reproduce faithfully: the closest, D\textsuperscript{2}MoE, has no
public implementation.

\begin{table}[!h]
\centering
\footnotesize
\setlength{\tabcolsep}{3pt}
\begin{tabular}{lrrrr}
\toprule
$\bar{k}$ & Uniform & Residual & Gate & DPA \\
          & -$k$    & -only    & -only & (ours) \\
\midrule
3 & 0.558387 & 0.514394 & 0.321453 & \textbf{0.306188} \\
4 & 0.261631 & 0.249736 & 0.145686 & \textbf{0.137787} \\
5 & 0.118282 & 0.113541 & 0.067866 & \textbf{0.064224} \\
6 & 0.041207 & 0.041108 & 0.030344 & \textbf{0.028952} \\
\bottomrule
\end{tabular}
\caption{Mean KL divergence on OLMoE-1B-7B-0125 with each allocation
signal isolated. Lower is better.}
\label{tab:signal-ablation}
\end{table}
Table~\ref{tab:signal-ablation} shows that DPA attains the lowest KLD at
every budget. Gate-only accounts for most of the gain over Uniform-$k$
($26.4$--$44.3\%$), while Residual-only contributes little
($0.2$--$7.9\%$): a static sensitivity profile is a weak signal on its
own. Combining the two still improves on Gate-only by $4.6$--$5.4\%$
everywhere, so the offline residual is complementary rather than
redundant --- as the objective's structure implies, the gate coefficient
scales the residual rather than replacing it.

\section{Discussion}

\paragraph{What does DPA-MoE show?}
DPA-MoE shows that, after the router selects the active experts, there
is still a meaningful precision-allocation problem. Uniform-$k$ assigns
the same chunk count to every routed expert, but routed experts differ
in two ways: their gate scores indicate different importance for the
current token, and their residual profiles indicate different
quantization sensitivity at each chunk count. DPA combines these two
signals into a gate-weighted residual objective, allowing the same
chunk budget to be used more effectively without modifying the
pretrained router or training an additional policy. The consistent
improvements over Uniform-$k$ suggest that expert precision should be
treated as a runtime allocation variable rather than a fixed property
of the encoded checkpoint.

\paragraph{How should the runtime results be interpreted?}
DPA-MoE improves the weight-size--quality frontier, but it is not a
free speedup at a fixed budget: at matched $\bar{k}$, per-dispatch
allocation and non-uniform chunk materialization lower steady-state TPS
and raise TTFT in our prototype. The benefit is clearer at matched
quality, since DPA often reaches a larger Uniform-$k$ budget's quality
with fewer chunks, recovering part of that overhead at a smaller
effective weight size. Its value is thus a better set of
weight-size--quality operating points from one encoded model, not
higher throughput at every fixed budget.

\paragraph{Where is DPA-MoE the right tool?}
For batched prefill $\mathcal{A}$ is the union of all tokens' selected
experts, so the $O(n^2\bar{k}k_{\mathrm{range}})$ allocator cost of
Section~\ref{sec:dp} grows with batch size while its benefit shrinks:
inference turns compute-bound and each expert is reused across more
tokens. DPA-MoE therefore suits memory-constrained, small-to-moderate
batches --- single-stream decoding, local serving --- and a deployment
should fall back to Uniform-$k$ past the batch size where allocation
stops paying for itself, a threshold set by the target device rather
than by the method.

\paragraph{Scope and extensions.}
SharedScale AnyBCQ is one prefix-stable encoding substrate for this
framework, not the only possible choice. The allocation method can be
applied to other progressive encodings as long as their precision units
can be treated as allocatable chunks with measurable residual profiles.
Likewise, DPA-MoE exposes a runtime precision interface that could be
used by higher-level policies such as phase-aware decoding, adaptive
quality control, or speculative decoding. This paper focuses on the
core allocation problem; optimized runtime integration and downstream
system composition are left as directions for future work.

\section{Conclusion}
We presented DPA-MoE, a runtime precision control framework for
any-precision MoE inference. Standard MoE routing determines which
experts are activated for each token, but leaves the selected experts
at a fixed precision. DPA-MoE adds a post-routing precision allocation
step: given the experts selected by the pretrained router and a fixed
chunk budget, it decides how many precision chunks each routed expert
activation should receive.

To support this formulation, we introduced SharedScale AnyBCQ, a
prefix-stable expert encoding that lets a single MoE checkpoint expose
multiple effective weight-memory budgets without retraining or
re-encoding. We then combined offline residual profiling with live
router gate scores to allocate chunks according to gate-weighted
reconstruction costs.

Experiments on four MoE models show that this allocation consistently
improves reconstruction fidelity over uniform chunk allocation and
yields a better overall weight-size--quality frontier. Runtime
measurements further show that these gains come with modest
steady-state throughput overhead in our current prototype.

Overall, DPA-MoE suggests that expert precision can be treated as a
runtime allocation variable rather than a fixed property of a deployed
checkpoint. By allowing a single encoded MoE checkpoint to trace a
fine-grained weight-size--quality frontier at runtime, DPA-MoE
establishes expert precision as a first-class inference-time
variable---one that downstream systems, adaptive controllers, and
deployment pipelines can optimize directly without maintaining
separate model variants.

\section{Limitations}
DPA-MoE optimizes the weight-size--quality frontier of a single encoded MoE checkpoint, but effective weight size is a proxy, not a complete latency model. Actual serving performance also depends on kernel fusion, batching, dispatch overhead, and hardware-specific scheduling, none of which are modeled here.

Our runtime is a lightweight prototype built on top of 
llama.cpp~\citep{ggerganov2023llama}. The allocator and variable chunk materialization are not fully fused with expert dispatch kernels, and no custom scheduler is used. The observed TTFT overhead and modest TPS reduction should therefore be interpreted as prototype costs rather than intrinsic costs of the method.

The allocation objective is a surrogate for downstream quality. DPA uses live router gate scores as a lightweight proxy for expert importance and offline residuals to estimate quality loss, but gate scores are imperfect and the objective ignores cross-expert residual interactions. Residual estimates are computed from finite calibration data and can contain small precision inversions, which we address via monotone projection. The dynamic program is therefore optimal for the projected surrogate, not necessarily for the true downstream loss.

Finally, our empirical scope covers four open-weight MoE models and a focused set of benchmarks, chosen to isolate the effect of non-uniform chunk allocation against a uniform baseline. They differ in scale, routing width, and shared-expert configuration, but we do not claim exhaustive generality across all MoE architectures, routing designs, hardware backends, or task domains. DPA-MoE also requires access to router gate scores, which may not be available in closed-weight deployments.

\section*{Ethical considerations}
All experiments use public model and dataset artifacts: Qwen3-30B-A3B,
DeepSeekMoE-16B-Chat, OLMoE-1B-7B-0125, Qwen1.5-MoE-A2.7B,
WikiText-2, C4/mC4, ARC, WinoGrande, and PIQA. We
did not collect new human-subject data, recruit annotators, or
redistribute model weights or dataset contents. Web-derived text used
for calibration and speed prompts may contain offensive content or
personally identifying strings inherited from the source corpora; we use
it only for aggregate evaluation and report only summary metrics.

AI assistants were used for coding support, documentation assistance,
and editorial checks. The authors reviewed the resulting code,
experiments, and manuscript text and are responsible for the correctness
and originality of the submission.

\section*{Acknowledgments}

This work was supported by Institute of Information \& Communications Technology Planning \& Evaluation (IITP) grants funded by the Korea government (MSIT) under Grant Nos. RS-2025-02218704 (Development of Chiplet-based DNN-SNN Hybrid Computing Processor for Large-Language-Models) and RS-2025-02218322 (Development of Semiconductor for Processing-In-Memory (PIM) Acceleration Using Multi-Hybrid Systems), each contributing 50\%.

\bibliography{reference}

\vspace{0.5\baselineskip}

\appendix

\section{Storage Overhead: AnyBCQ vs.\ SharedScale AnyBCQ}
\label{app:memory-analysis}

To support any-precision inference across the full 2--8 bit range 
at runtime, both AnyBCQ and SharedScale AnyBCQ must store all 
bitplanes and their associated scale parameters simultaneously. 
Although the two encodings require the same eight bitplanes, they 
differ substantially in how many scale sets must be retained.

In AnyBCQ, scales are jointly re-optimized at every precision tier, 
so each tier maintains its own independent scale set. Supporting 
the full 2--8 bit range therefore requires storing 
$2 + 3 + \cdots + 8 = 35$ scale sets in total. In SharedScale 
AnyBCQ, a single scale set is fixed per bitplane and shared across 
all tiers, so only 8 scale sets are needed regardless of the 
supported precision range.

\begin{table}[h]
\centering
\begin{tabular}{lrrrr}
\toprule
Enc. & Bits & Scale & Total & \% \\
\midrule
AnyBCQ & 4,608 & 2,520 & 7,128 & 35.4 \\
SS-AnyBCQ & 4,608 & 576 & 5,184 & 11.1 \\
\bottomrule
\end{tabular}
\caption{Per-expert storage breakdown in KB for full 2--8 bit 
support on Qwen3-30B-A3B.}
\label{tab:storage-comparison}
\end{table}

Table~\ref{tab:storage-comparison} quantifies this difference for 
a single expert in Qwen3-30B-A3B (hidden=2048, intermediate=768, 
group size=128, Up+Gate+Down matrices). Each bitplane occupies 
576~KB and each scale set occupies 72~KB under this architecture.

The scale overhead in AnyBCQ is 4.4$\times$ larger than in 
SharedScale AnyBCQ, increasing total per-expert storage by 37.5\%. 
More critically, this asymmetry compounds at runtime. In DPA-MoE, 
expert precision is reallocated on every token where routing 
confidence changes. Each upward precision transition in AnyBCQ 
requires materializing all accumulated scale sets for the target 
tier, since they are not shared across tiers. In SharedScale 
AnyBCQ, only a single new bitplane and its associated scale set 
must be materialized per transition, regardless of the current 
precision level. The storage savings in Table~\ref{tab:storage-comparison} 
therefore translate directly into reduced materialization overhead 
per transition, which accumulates across the full generation 
sequence.

\section{Cross-Term Analysis}
\label{app:cross-term}
To validate the diagonal approximation used in 
Section~\ref{sec:allocation}, we measure the relationship 
between the exact squared reconstruction error and its 
diagonal approximation at each precision level $k$.
For each MoE dispatch, we compute:
\begin{equation}
\begin{aligned}
\text{exact}_k &= \left\|\sum_e g_e \cdot
\Delta E_e^{(k)}\right\|^2, \\
\text{diag}_k &= \sum_e g_e^2 \cdot
\left\|\Delta E_e^{(k)}\right\|^2,
\end{aligned}
\label{eq:cross-term-defs}
\end{equation}
and report their ratio, the relative cross-term 
contribution, and rank correlations across dispatches, 
averaged over 10 sequences of 4{,}096 tokens from the 
WikiText-2 test set on Qwen3-30B-A3B (Table~\ref{tab:cross-term}).

\begin{table}[h]
\centering
\begin{tabular*}{\columnwidth}{@{\extracolsep{\fill}}ccccc}
\toprule
$k$ & diag/ex. & cross/ex. & Pearson & Spearman \\
\midrule
2 & 0.402 & 0.598 & 0.987 & 0.999 \\
3 & 0.881 & 0.119 & 0.972 & 1.000 \\
4 & 0.916 & 0.084 & 0.980 & 1.000 \\
5 & 0.733 & 0.268 & 0.972 & 1.000 \\
6 & 0.831 & 0.169 & 0.955 & 1.000 \\
7 & 0.871 & 0.129 & 0.979 & 1.000 \\
8 & 1.000 & 0.000 & 0.999 & 1.000 \\
\bottomrule
\end{tabular*}
\caption{Cross-term analysis per precision level $k$.}
\label{tab:cross-term}
\end{table}

At $k{=}2$, cross terms contribute substantially ($59.8\%$),
but this precision level represents the minimum chunk count
to which every expert defaults, so no allocation decision
is made in this regime.

For $k{\geq}3$, cross terms contribute at most $27\%$ of 
the true error. Critically, the Spearman rank correlation 
between the exact and diagonal objectives exceeds $0.999$ 
across all precision levels and Pearson exceeds $0.95$. 
Since the knapsack allocator optimizes relative costs 
across experts rather than absolute magnitudes, the 
near-perfect rank preservation confirms that the diagonal 
approximation reliably guides allocation toward 
near-optimal chunk assignments in practice. Separability also makes the
allocation a knapsack problem at all: with cross terms, one expert's
cost would depend on every other expert's assignment.

\section{Per-Task Benchmark Results}
\label{app:pertask}

\begin{figure*}[!p]
  \centering
  \begin{tabular}{cc}
    \includegraphics[width=0.47\textwidth]{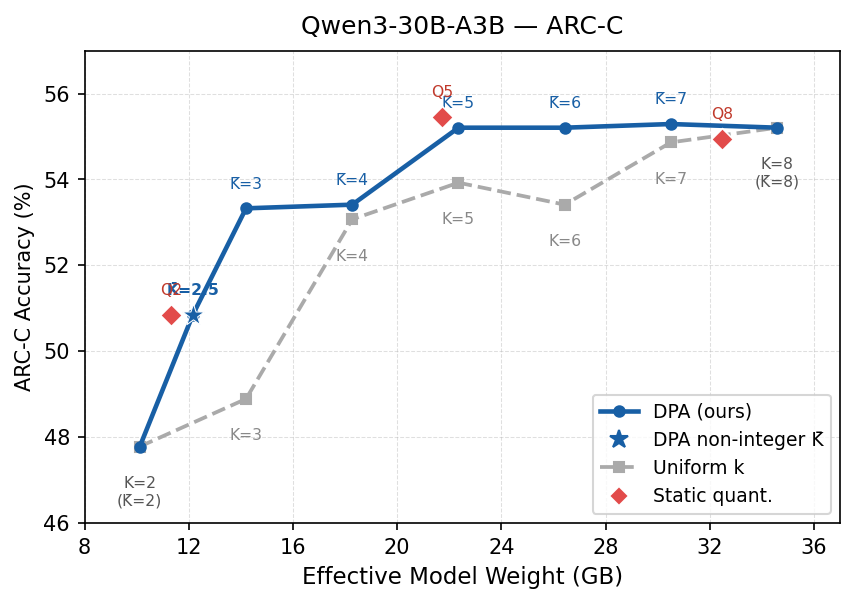} &
    \includegraphics[width=0.47\textwidth]{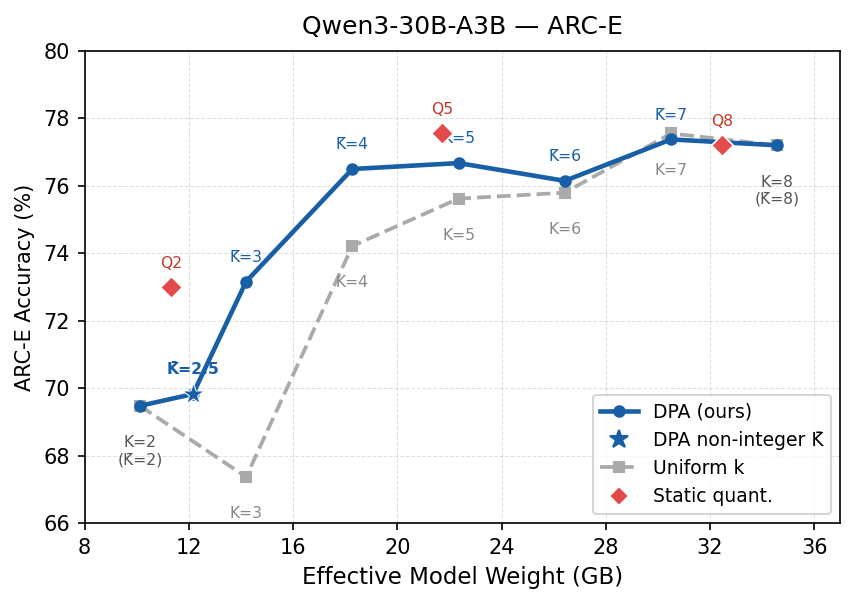} \\[6pt]
    \includegraphics[width=0.47\textwidth]{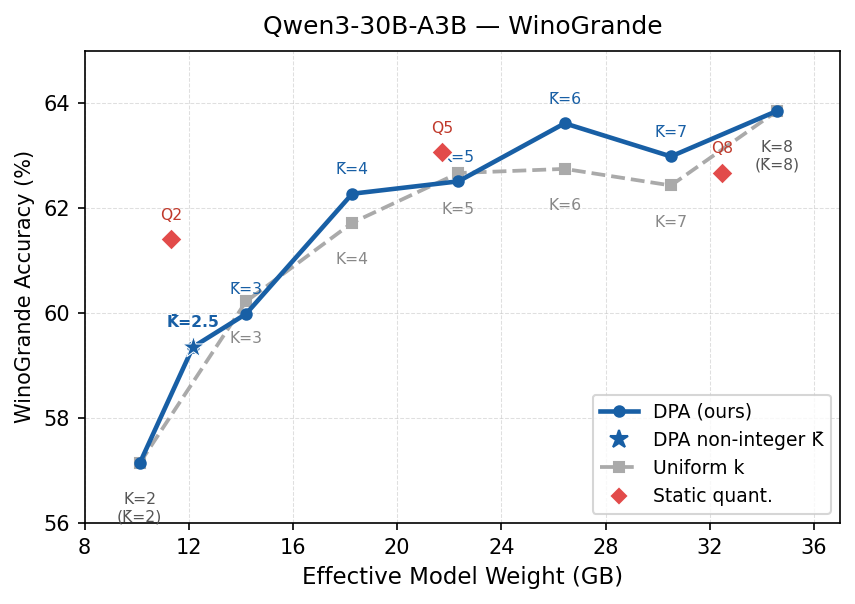} &
    \includegraphics[width=0.47\textwidth]{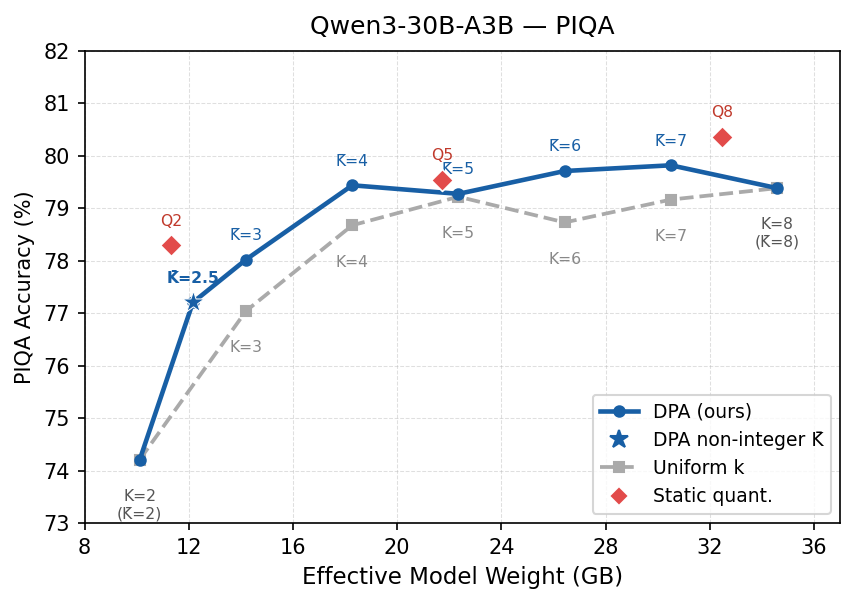} \\
  \end{tabular}
  \caption{Per-task benchmark results for Qwen3-30B-A3B.}
  \label{fig:qwen3_pertask}
\end{figure*}

\begin{figure*}[!p]
  \centering
  \begin{tabular}{cc}
    \includegraphics[width=0.47\textwidth]{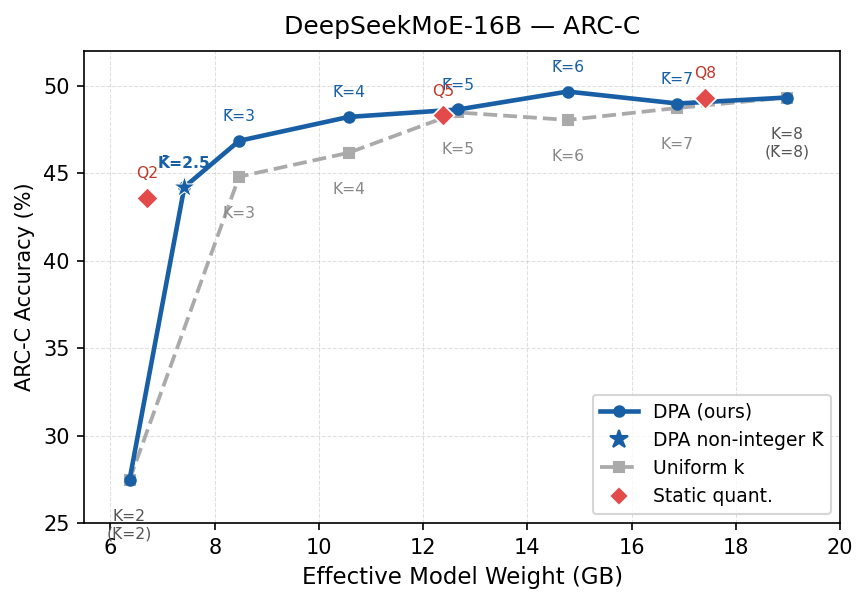} &
    \includegraphics[width=0.47\textwidth]{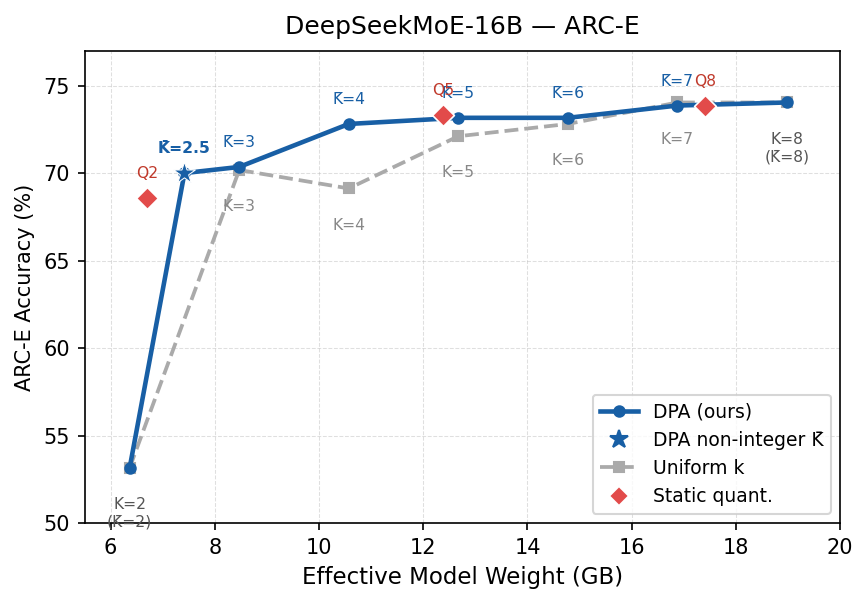} \\[6pt]
    \includegraphics[width=0.47\textwidth]{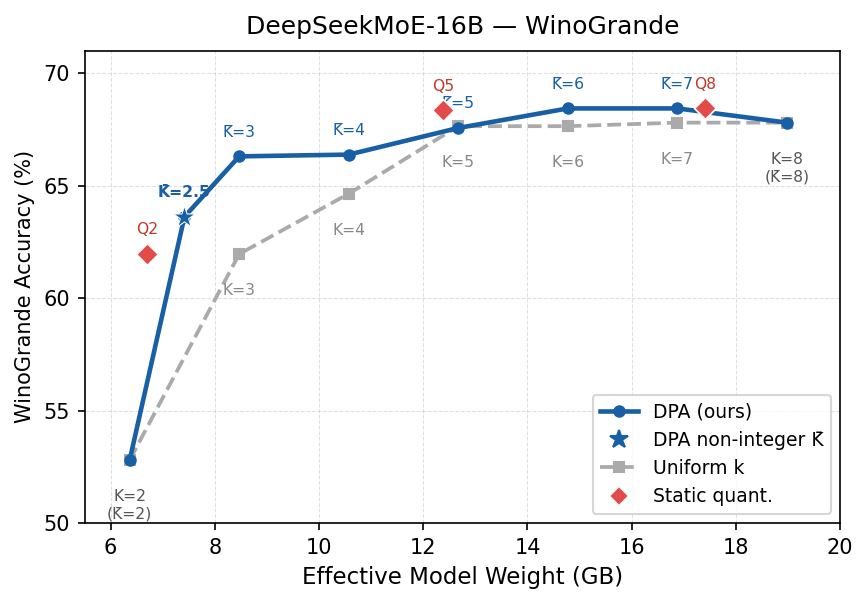} &
    \includegraphics[width=0.47\textwidth]{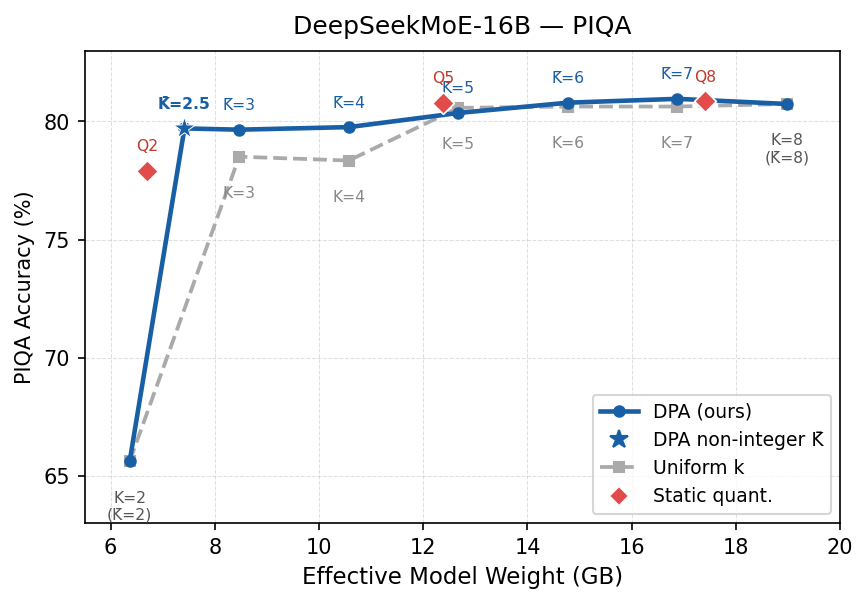} \\
  \end{tabular}
  \caption{Per-task benchmark results for DeepSeekMoE-16B-Chat.}
  \label{fig:deepseek_pertask}
\end{figure*}

Figures~\ref{fig:qwen3_pertask} and~\ref{fig:deepseek_pertask}
report per-task accuracy across ARC-Challenge, ARC-Easy,
WinoGrande, and PIQA for Qwen3-30B-A3B and DeepSeekMoE-16B-Chat
respectively. DPA consistently outperforms Uniform~$k$ at
matched memory budgets across all four tasks on both models,
with the largest gains at aggressive compression.

\begin{table*}[!t]
\centering
\small
\begin{tabular*}{\textwidth}{@{\extracolsep{\fill}}lccccc}
\toprule
\textbf{Method} & \textbf{Training-free} & \textbf{Router intact}
& \textbf{Token-aware} & \textbf{Fine-grained $k$}
& \makecell{\textbf{No re-fetch}\\\textbf{on upgrade}} \\
\midrule
MxMoE~\citep{duanmu2025mxmoe}                 & \cmark & \cmark & \xmark & \xmark & \xmark \\
HOBBIT~\citep{tang2024hobbit}                 & \cmark & \cmark & \pmark & \xmark & \xmark \\
DynaExq~\citep{chu2025dynaexq}                & \cmark & \cmark & \xmark & \xmark & \xmark \\
D\textsuperscript{2}MoE~\citep{wang2025d2moe} & \xmark & \cmark & \cmark & \xmark & \cmark \\
\midrule
\textbf{DPA-MoE (ours)}                       & \cmark & \cmark & \cmark & \cmark & \cmark \\
\bottomrule
\end{tabular*}
\caption{Comparison with existing dynamic and mixed expert-precision
methods. \pmark{}: HOBBIT switches precision only for cache-miss
experts, so the decision is gated by cache state rather than by the gate
score itself.}
\label{tab:method-comparison}
\end{table*}
\begin{table*}[t]
\centering
\small
\begin{tabular*}{\textwidth}{@{\extracolsep{\fill}}lrrrrr@{\hskip 1.5em}lrrrrr}
\toprule
\multicolumn{6}{c}{\textbf{OLMoE-1B-7B-0125}}
& \multicolumn{6}{c}{\textbf{Qwen1.5-MoE-A2.7B}} \\
\cmidrule(lr){1-6}\cmidrule(lr){7-12}
& & \multicolumn{2}{c}{Mean KLD} & \multicolumn{2}{c}{Avg Acc (\%)}
& & & \multicolumn{2}{c}{Mean KLD} & \multicolumn{2}{c}{Avg Acc (\%)} \\
\cmidrule(lr){3-4}\cmidrule(lr){5-6}\cmidrule(lr){9-10}\cmidrule(lr){11-12}
Config & GB & Uni & DPA & Uni & DPA
& Config & GB & Uni & DPA & Uni & DPA \\
\midrule
$k{=}2$ & 2.42 & 1.06908 & 1.06908 & 54.63 & 54.63
& $k{=}2$ &  6.30 & 0.25540 & 0.25540 & 60.98 & 60.98 \\
$k{=}3$ & 3.33 & 0.56371 & \textbf{0.32776} & 60.60 & \textbf{62.40}
& $k{=}3$ &  7.93 & 0.10721 & \textbf{0.04213} & 62.60 & \textbf{62.78} \\
$k{=}4$ & 4.23 & 0.26034 & \textbf{0.14780} & 63.34 & \textbf{64.17}
& $k{=}4$ &  9.56 & 0.05210 & \textbf{0.01630} & 62.04 & \textbf{63.45} \\
$k{=}5$ & 5.14 & 0.11913 & \textbf{0.06908} & 64.52 & \textbf{64.67}
& $k{=}5$ & 11.19 & 0.01953 & \textbf{0.00694} & 62.63 & \textbf{63.52} \\
$k{=}6$ & 6.05 & 0.04293 & \textbf{0.03166} & \textbf{64.88} & 64.57
& $k{=}6$ & 12.83 & 0.00611 & \textbf{0.00323} & 63.66 & \textbf{64.17} \\
$k{=}7$ & 6.95 & 0.02145 & \textbf{0.01498} & 64.62 & \textbf{65.07}
& $k{=}7$ & 14.46 & 0.00323 & \textbf{0.00187} & 63.70 & \textbf{64.30} \\
$k{=}8$ & 7.86 & 0.01048 & 0.01048 & 65.39 & 65.39
& $k{=}8$ & 16.09 & 0.00148 & 0.00148 & 64.46 & 64.46 \\
\midrule
Q2\_K    & 2.58 & \multicolumn{2}{c}{0.600} & \multicolumn{2}{c}{61.66}
& Q2\_K    &  5.49 & \multicolumn{2}{c}{0.288} & \multicolumn{2}{c}{58.97} \\
Q5\_K\_M & 4.93 & \multicolumn{2}{c}{0.026} & \multicolumn{2}{c}{64.58}
& Q5\_K\_M & 10.09 & \multicolumn{2}{c}{0.008} & \multicolumn{2}{c}{63.88} \\
Q8\_0    & 7.36 & \multicolumn{2}{c}{0.006} & \multicolumn{2}{c}{64.52}
& Q8\_0    & 14.18 & \multicolumn{2}{c}{0.001} & \multicolumn{2}{c}{64.31} \\
\bottomrule
\end{tabular*}
\caption{Mean KL divergence and macro-average benchmark accuracy versus
effective weight-memory budget on the two additional MoE architectures.}
\label{tab:extra-models}
\end{table*}

\section{Comparison with Dynamic Expert-Precision Methods}
\label{app:method-comparison}

Table~\ref{tab:method-comparison} expands the positioning discussion in
the Related Work section into a per-axis comparison.

MxMoE assigns mixed precision per expert, but the allocation is fixed
before deployment, is independent of the current input, and does not
support runtime switching. HOBBIT selects a substitute precision for
cache-miss experts only; the mechanism is gated by cache state rather
than by the gate score, and it does not apply to cache-hit experts.
DynaExq promotes and demotes experts between high- and low-precision
resident sets using long-horizon expert hotness estimated from router
traces, so it relies on aggregate statistics rather than each expert's
importance to the current token. D\textsuperscript{2}MoE uses a nested
Matryoshka quantization in which higher bit-widths encapsulate lower
ones, so upgrading precision reuses already-loaded bits --- a property
it shares with our prefix-decodable design. It nonetheless supports
only discrete tiers, offers no non-integer interpolation, and requires
training a separate per-expert plug-in network to decide bit-width.
DPA-MoE is, to our knowledge, the only method that is simultaneously
training-free, token-aware, and fine-grained in $k$ while leaving the
router untouched.

We do not report a head-to-head empirical comparison against these
methods. D\textsuperscript{2}MoE is the closest in capability, but we
are not aware of a public implementation, and its per-expert plug-in
networks would have to be trained before any comparison could be run.
The main comparison in the paper is therefore against Uniform-$k$, and
the signal ablation of Section~\ref{sec:signal-ablation} isolates the
contribution of each allocation signal in place of a cross-method
baseline.

\section{Additional MoE Architectures}
\label{app:extra-models}

Table~\ref{tab:extra-models} gives the full budget sweep behind
Section~\ref{sec:generality} for OLMoE-1B-7B-0125 and
Qwen1.5-MoE-A2.7B, using the same protocol as the two models in the
main text. DPA and Uniform-$k$ necessarily coincide at
$\bar{k}{=}k_{\min}{=}2$ and $\bar{k}{=}k_{\max}{=}8$, where the budget
admits only one feasible allocation. The static quantization rows are
reference points at the nearest bit-width, not matched-budget
comparisons.

The reduction in KL divergence is consistent across the interior of the
sweep on both models. On OLMoE the mean KLD falls from $0.564$ to
$0.328$ at $\bar{k}{=}3$, from $0.260$ to $0.148$ at $\bar{k}{=}4$, and
from $0.119$ to $0.069$ at $\bar{k}{=}5$; on Qwen1.5-MoE it falls from
$0.107$ to $0.042$, from $0.052$ to $0.016$, and from $0.020$ to
$0.007$ at the same three budgets. Downstream accuracy improves at most
matched budgets, the largest gains being $+1.80$ points at
$\bar{k}{=}3$ on OLMoE and $+1.41$ points at $\bar{k}{=}4$ on
Qwen1.5-MoE. The one budget where Uniform-$k$ is ahead is
$\bar{k}{=}6$ on OLMoE, by $0.31$ points, even though DPA attains the
lower KLD there.

The two models were chosen to differ from the main-text pair on the
axes most likely to interact with post-routing allocation: the presence
of always-on shared experts, of which OLMoE has none and Qwen1.5-MoE
has four, and routing scale, $8$-of-$64$ against $4$-of-$60$ plus four
shared. Together with Qwen3-30B-A3B and DeepSeekMoE-16B-Chat, the four
models cover both settings on each axis, which is the basis for the
generality claim in Section~\ref{sec:generality}.

\section{Composing with Dynamic Top-$k$ Routing}
\label{app:dynamic-topk}

The experiments in the main text assume a fixed top-$k$ active expert
count. The budget $B=\mathrm{round}(|\mathcal{A}|\cdot\bar{k})$ of
Section~\ref{sec:dp} is however already defined with respect to whatever
$|\mathcal{A}|$ the router produces at each dispatch, so nothing in the
formulation requires $|\mathcal{A}|$ to be constant. We verify this
directly. Following~\citet{huang-etal-2024-harder}, we emulate
dynamic-$k$ routing by applying a nucleus~\citep{holtzman2020curious}
cutoff to the already-computed expert distribution of
OLMoE-1B-7B-0125-Instruct at inference time. The router itself is
untouched and no retraining is performed; the cumulative-probability
threshold is applied post hoc. At $p{=}0.25$ the router activates $4.0$
experts per token on average.

\paragraph{DPA composes with dynamic routing.}
Table~\ref{tab:dynamic-topk} compares DPA against Uniform-$k$ on the
same routing trajectory and chunk budget. DPA improves PIQA accuracy by
$+0.37$ percentage points on average across the budget sweep, without
any retraining or routing-specific loss design. The gain is not uniform
--- it is largest at $\bar{k}{=}6$ and $\bar{k}{=}7$ and slightly
negative at $\bar{k}{=}5$ --- but the method transfers without
modification.

\begin{table}[H]
\centering
\footnotesize
\setlength{\tabcolsep}{3.5pt}
\renewcommand{\arraystretch}{0.92}
\begin{tabular}{lrrr}
\toprule
$\bar{k}$ & DPA (\%) & Uniform-$k$ (\%) & $\Delta$ \\
\midrule
3 & \textbf{72.48} & 72.22 & $+0.26$ \\
4 & \textbf{73.10} & 73.08 & $+0.02$ \\
5 & 73.44 & \textbf{73.54} & $-0.10$ \\
6 & \textbf{74.45} & 73.78 & $+0.67$ \\
7 & \textbf{74.23} & 73.25 & $+0.98$ \\
\midrule
Avg & \textbf{73.54} & 73.17 & $+0.37$ \\
\bottomrule
\end{tabular}
\caption{PIQA accuracy under dynamic top-$p$ routing ($p{=}0.25$,
average $4.0$ active experts per token) on OLMoE-1B-7B-0125-Instruct,
averaged over 5 runs.}
\label{tab:dynamic-topk}
\vspace{0.50em}

\begin{tabular}{llrr}
\toprule
Series & Point & MB/token & PIQA (\%) \\
\midrule
\multirow{3}{*}{top-$p$ + Q8\_0}
 & $p{=}0.13$ & 213.9 & 61.92 \\
 & $p{=}0.25$ & 427.8 & 74.27 \\
 & $p{=}0.35$ & 631.1 & 76.44 \\
\midrule
\multirow{3}{*}{top-8 + DPA}
 & $\bar{k}{=}2$ & 239.1 & 70.99 \\
 & $\bar{k}{=}4$ & 465.6 & 77.49 \\
 & $\bar{k}{=}6$ & 692.1 & 78.30 \\
\bottomrule
\end{tabular}
\caption{Accuracy at matched memory read per token on
OLMoE-1B-7B-0125-Instruct. The top-$p$ rows activate on average
$2.0$, $4.0$, and $6.0$ experts respectively. Reducing the number of
routed experts and reducing per-expert precision are two ways to spend
the same byte budget.}
\label{tab:memory-matched}
\end{table}

\paragraph{Precision and expert count are not interchangeable.}
Dynamic top-$k$ routing and DPA both reduce the bytes read per token, so
comparing them at matched memory read is a fair contrast.
Table~\ref{tab:memory-matched} sweeps both knobs. At the two closest
points --- top-$p$ activating $4.0$ experts at 8-bit
($427.8$~MB/token) and static top-8 routing with DPA at $\bar{k}{=}4$
($465.6$~MB/token) --- DPA reads $8.8\%$ more and scores $3.22$ points
higher. The ordering holds across the compression range we swept: at
matched memory read, shortening every selected expert loses less than
dropping experts wholesale.

\paragraph{The two axes compose.}
Under fixed top-8 routing, DPA already traverses the size--quality curve
from $k_{\min}{=}2$ to $K{=}8$ from a single encoding. Adding dynamic
routing turns the previously fixed active expert count into a second
adjustable axis, giving a two-dimensional operating range over
$(|\mathcal{A}|, \bar{k})$ --- from a minimal-compute point
$(|\mathcal{A}|_{\min}, k_{\min})$ to the full-fidelity point
$(|\mathcal{A}|_{\max}, K)$ --- still from one artifact.

\makeatletter
\setlength{\@fptop}{0pt}
\setlength{\@fpsep}{20pt}
\setlength{\@fpbot}{0pt plus 1fil}
\makeatother
\setcounter{dbltopnumber}{3}

\section{Detailed Experiment Configuration}
\label{app:experiment-config}

Tables~\ref{tab:model-config}--\ref{tab:speed-config} summarize the
reproducibility-critical settings used in the paper.
Table~\ref{tab:model-config} lists which checkpoints were encoded and
on what hardware, Table~\ref{tab:encoding-config} how they were encoded
and profiled, and Tables~\ref{tab:eval-config}
and~\ref{tab:speed-config} how the resulting artifacts were measured
for quality and for speed. We identify models by public Hugging Face
repository IDs and source revision hashes, and report only the budgets
and runtime knobs that affect the reported quality and speed numbers;
all random seeds are fixed to 0 unless stated otherwise. The
operating-point rows list the full sweeps that were run, not only the
points quoted in the main text.

\section{Artifact, License, and Reproducibility Notes}
\label{app:artifact-license}

Tables~\ref{tab:model-artifacts} and~\ref{tab:data-software-artifacts}
summarize the external artifacts used in the experiments, their public
source identifiers, and how they are used in this work. Model weights,
datasets, and generated outputs are not redistributed with the
supplementary package; the package contains source code, scripts, and
configuration files. Users who reproduce the experiments should obtain
external artifacts from their original sources and follow the
corresponding model-card or dataset-card terms.

The reported experiments use only public model and dataset artifacts.
For web-derived text sources such as C4 and mC4, we use the corpora
only as prompts or calibration text and report aggregate metrics. We do
not publish raw sampled prompts, generated answers, or any newly
collected human-subject data.

\paragraph{What each artifact is needed for.}
Reproducing the encoded weights needs only the base checkpoints of
Table~\ref{tab:model-artifacts} and the encoder settings of
Table~\ref{tab:encoding-config}. The static GGUF rows serve the
fixed-precision baselines, and the Instruct checkpoint only the
dynamic top-$k$ study of Appendix~\ref{app:dynamic-topk}.

\begin{table*}[!t]
\centering
\footnotesize
\setlength{\tabcolsep}{3pt}
\renewcommand{\arraystretch}{0.95}
\begin{tabular}{P{0.13\textwidth}P{0.18\textwidth}P{0.61\textwidth}}
\toprule
\textbf{Scope} & \textbf{Field} & \textbf{Value} \\
\midrule
\multirow{4}{*}{Models}
& Qwen3-30B-A3B & 48 layers; top-8 routing over 128 experts. \\
& DeepSeekMoE-16B-Chat & 28 layers; top-6 routing over 64 routed experts plus 2 shared experts. \\
& OLMoE-1B-7B-0125 & 16 layers; top-8 routing over 64 experts; no shared experts. \\
& Qwen1.5-MoE-A2.7B & 24 layers; top-4 routing over 60 routed experts plus 4 shared experts. \\
\midrule
\multirow{4}{*}{Artifacts}
& Base HF checkpoints & \texttt{Qwen/Qwen3-30B-A3B}@\texttt{ad44e77}; \texttt{deepseek-ai/deepseek-moe-16b-chat}@\texttt{eefd8ac}; \texttt{allenai/OLMoE-1B-7B-0125}@\texttt{9b0c1aa}; \texttt{Qwen/Qwen1.5-MoE-A2.7B}@\texttt{1a758c5}. \\
& Instruct checkpoint & \texttt{allenai/OLMoE-1B-7B-0125-Instruct}@\texttt{b89a7c4}; used only for the dynamic top-$k$ study of Appendix~\ref{app:dynamic-topk}. \\
& Static GGUF artifacts & \texttt{unsloth/Qwen3-30B-A3B-GGUF}@\texttt{d5b1d57}; \texttt{mradermacher/deepseek-moe-16b-chat-GGUF}@\texttt{fe95dea}; the OLMoE-1B-7B-0125 and Qwen1.5-MoE-A2.7B artifacts are converted locally from the base checkpoints with llama.cpp. \\
& DPA encoded weights & Generated from the base HF checkpoints above; no separate pretrained checkpoint is used. \\
\midrule
\multirow{2}{*}{Compute}
& Hardware & NVIDIA RTX PRO 6000 Blackwell Server Edition GPUs; 96~GiB VRAM per GPU; 1228~GiB host memory. \\
& Budget & Below 1{,}000 GPU-hours across encoding, profiling, KLD, benchmark, and speed experiments. \\
\bottomrule
\end{tabular}
\caption{Models, source artifacts, and compute.}
\label{tab:model-config}
\end{table*}

\begin{table*}[!t]
\centering
\footnotesize
\setlength{\tabcolsep}{3pt}
\renewcommand{\arraystretch}{0.95}
\begin{tabular}{P{0.13\textwidth}P{0.18\textwidth}P{0.61\textwidth}}
\toprule
\textbf{Scope} & \textbf{Field} & \textbf{Value} \\
\midrule
\multirow{7}{*}{Encoding}
& Target tensors & Expert FFN weights only; non-expert tensors use RTN Q8\_0. \\
& Encoding & SharedScale AnyBCQ with prefix-decodable chunks. \\
& Chunk range & $k_{\min}{=}2$, $K{=}8$. \\
& Quantization shape & Group size 128; asymmetric reconstruction; learned $\beta$; no any-precision $\alpha$. \\
& Optimization & FP32 reconstruction; Adam learning rates $10^{-4}/10^{-4}/10^{-4}$; 20 optimization iterations. \\
& Initialization & 20 rounds of alternating refinement per BCQ sign bitplane. \\
& Calibration & 512 samples; batch size 2; sequence length 2048; WikiText-2 training text plus held-out calibration text. \\
\midrule
\multirow{5}{*}{Profiling}
& Target & Per-layer, per-expert, per-chunk residual lookup table. \\
& Data & WikiText-2 training text. \\
& Length & Sequence length 2048. \\
& Limits & Maximum 16{,}384 profiled tokens; maximum 256 samples per expert. \\
& Postprocess & Monotone-projected LUT for reported Qwen3 KLD, benchmark, and speed runs. \\
\bottomrule
\end{tabular}
\caption{Encoding and residual-profiling configuration.}
\label{tab:encoding-config}
\end{table*}

\begin{table*}[!p]
\centering
\footnotesize
\setlength{\tabcolsep}{3pt}
\renewcommand{\arraystretch}{0.95}
\begin{tabular}{P{0.13\textwidth}P{0.18\textwidth}P{0.61\textwidth}}
\toprule
\textbf{Scope} & \textbf{Field} & \textbf{Value} \\
\midrule
\multirow{6}{*}{KLD}
& Metric & Mean KL divergence against the BF16 reference. \\
& Dataset & WikiText-2 raw test text. \\
& Input & 10 sequences; 4096 tokens per sequence. \\
& Runtime & Context 4096; batch 1; microbatch 1. \\
& Operating points & Static Q2\_K/Q5\_K\_M/Q8\_0; Uniform $k\in\{2,3,4,5,6,7,8\}$; DPA $\bar{k}\in\{2.5,3,3.5,4,4.5,5,5.5,6,6.5,7\}$. \\
& Execution & Full pinned expert chunks; CUDA capture where available; VRAM caps: 40~GB for Qwen3, 30~GB for DeepSeekMoE. \\
\midrule
\multirow{6}{*}{Benchmark}
& Metric & Multiple-choice accuracy. \\
& Tasks & ARC-Challenge, ARC-Easy, PIQA, and WinoGrande. \\
& Runtime & llama.cpp multiple-choice perplexity path; context 4096; batch 2048; microbatch 512. \\
& Decoding controls & One parallel request; temperature 0; reasoning budget 0. \\
& Operating points & Static Q2\_K/Q5\_K\_M/Q8\_0; Uniform $k\in\{2,3,4,5,6,7,8\}$; DPA $\bar{k}\in\{2.5,3,4,5,6,6.5,7\}$. \\
& Execution & Full pinned expert chunks; CUDA capture where available; VRAM caps match KLD. \\
\bottomrule
\end{tabular}
\caption{Reconstruction-fidelity and downstream benchmark configuration.}
\label{tab:eval-config}
\end{table*}

\begin{table*}[!p]
\centering
\footnotesize
\setlength{\tabcolsep}{3pt}
\renewcommand{\arraystretch}{0.95}
\begin{tabular}{P{0.13\textwidth}P{0.18\textwidth}P{0.61\textwidth}}
\toprule
\textbf{Scope} & \textbf{Field} & \textbf{Value} \\
\midrule
\multirow{5}{*}{Workload}
& Metrics & Generation TPS and time-to-first-token. \\
& Datasets & WikiText-2, C4-en, and C4-ko. \\
& Samples & 100 sampled prompts per dataset. \\
& Input/output & 1024-token prompts; 256 generated tokens. \\
& Repetitions & 6 warm-up runs and 1 measured run per sample. \\
\midrule
\multirow{5}{*}{Runtime}
& Batching & Context 4096; batch 2048; microbatch 256. \\
& Decoding controls & Greedy decoding; EOS ignored; prompt cache disabled. \\
& Operating points & Uniform $k\in\{3,4,5,6,7\}$ and DPA $\bar{k}\in\{3,4,5,6,7\}$. \\
& Execution & Pinned expert chunks; 40~GB VRAM cap; no host cache; CUDA capture disabled. \\
& Kernels & Optimized mixed-mode kernels with GPU planner enabled. \\
\bottomrule
\end{tabular}
\caption{Inference-speed measurement configuration.}
\label{tab:speed-config}
\end{table*}

\begin{table*}[!p]
\centering
\footnotesize
\setlength{\tabcolsep}{4pt}
\begin{tabular}{P{0.19\textwidth}P{0.25\textwidth}P{0.18\textwidth}P{0.30\textwidth}}
\toprule
\textbf{Artifact} & \textbf{Source / Version} & \textbf{Terms} & \textbf{Use in this paper} \\
\midrule
Qwen3-30B-A3B & \texttt{Qwen/\allowbreak Qwen3-30B-\allowbreak A3B@\allowbreak ad44e77} & Model-card terms; Apache-2.0 at access. & Base MoE checkpoint used for encoding and evaluation; weights are not redistributed. \\
Qwen3 GGUF & \texttt{unsloth/\allowbreak Qwen3-30B-A3B-\allowbreak GGUF@\allowbreak d5b1d57} & Inherited model-card terms. & Static Q2\_K/Q5\_K\_M/Q8\_0 comparison; artifacts are downloaded separately. \\
DeepSeekMoE-16B & \texttt{deepseek-ai/\allowbreak deepseek-moe-\allowbreak 16b-chat@\allowbreak eefd8ac} & DeepSeek model-card terms. & Base MoE checkpoint used for encoding and evaluation; weights are not redistributed. \\
DeepSeekMoE GGUF & \texttt{mradermacher/\allowbreak deepseek-moe-\allowbreak 16b-chat-GGUF@\allowbreak fe95dea} & Inherited model-card terms. & Static Q2\_K/Q5\_K\_M/Q8\_0 comparison; artifacts are downloaded separately. \\
OLMoE-1B-7B & \texttt{allenai/\allowbreak OLMoE-1B-7B-0125@\allowbreak 9b0c1aa} & Model-card terms at access. & Base MoE checkpoint for the additional-architecture sweep; weights are not redistributed. \\
OLMoE-1B-7B Instruct & \texttt{allenai/\allowbreak OLMoE-1B-7B-0125-\allowbreak Instruct@\allowbreak b89a7c4} & Model-card terms at access. & Router trajectory for the dynamic top-$k$ study; weights are not redistributed. \\
Qwen1.5-MoE-A2.7B & \texttt{Qwen/\allowbreak Qwen1.5-MoE-A2.7B@\allowbreak 1a758c5} & Model-card terms at access. & Base MoE checkpoint for the additional-architecture sweep; weights are not redistributed. \\
OLMoE / Qwen1.5-MoE GGUF & Converted locally from the base checkpoints with llama.cpp & Inherited model-card terms. & Static Q2\_K/Q5\_K\_M/Q8\_0 comparison; produced from the base checkpoints listed above, not redistributed. \\
\bottomrule
\end{tabular}
\caption{Model artifacts, versions, terms, and intended use.}
\label{tab:model-artifacts}
\end{table*}

\begin{table*}[!p]
\centering
\footnotesize
\setlength{\tabcolsep}{4pt}
\begin{tabular}{P{0.19\textwidth}P{0.25\textwidth}P{0.18\textwidth}P{0.30\textwidth}}
\toprule
\textbf{Artifact} & \textbf{Source / Version} & \textbf{Terms} & \textbf{Use in this paper} \\
\midrule
WikiText-2 & Raw train/test splits \citep{merity2017pointer} & Dataset/source terms. & Calibration, profiling, KLD, and speed prompts. \\
C4 / mC4 & C4-en \citep{raffel2020t5}; Korean mC4 \citep{xue2021mt5} & Web-derived dataset/source terms. & Speed prompts only; possible inherited web-text risks are discussed in Ethical considerations. \\
ARC, PIQA, WinoGrande & Standard evaluation splits \citep{clark2018think,bisk2020piqa,sakaguchi2021winogrande} & Dataset/source terms. & Multiple-choice downstream evaluation; no new labels or human-subject collection. \\
llama.cpp & ggml-org/\allowbreak llama.cpp \citep{ggerganov2023llama} & MIT. & Prototype runtime substrate; upstream notices are retained in the supplementary package. \\
AnyBCQ & Reference code \citep{anybcq2025} & Apache-2.0. & Encoding substrate and kernel reference; upstream notices are retained in the supplementary package. \\
\bottomrule
\end{tabular}
\caption{Dataset and software artifacts, terms, and intended use.}
\label{tab:data-software-artifacts}
\end{table*}

\end{document}